\documentclass[pdflatex,sn-mathphys-num,twoside,iicol]{sn-jnl}

\usepackage{geometry}
\usepackage{graphicx}%
\usepackage{multirow}%
\usepackage{amsmath,amssymb,amsfonts}%
\usepackage{amsthm}%
\usepackage{mathrsfs}%
\usepackage[title]{appendix}%
\usepackage{xcolor}%
\usepackage{textcomp}%
\usepackage{manyfoot}%
\usepackage{booktabs}%
\usepackage{algorithm}%
\usepackage{algorithmicx}%
\usepackage{algpseudocode}%
\usepackage{listings}%
\usepackage{enumitem}

\theoremstyle{thmstyleone}%
\theoremstyle{thmstyletwo}%

\theoremstyle{thmstylethree}%

\begin{document}

\title[Article Title]{PatchAdaNet: Unsupervised class incremental learning using local patches and global adapters}



\author[1,2]{\fnm{Kamel} \sur{Aouaidjia}}\email{kamel@henu.edu.cn}

\author[1,2]{\fnm{Yanchao} \sur{Duan}}\email{yanchao.duan@foxmail.com}
\author[1,2]{\fnm{Zihui} \sur{Zhen}}\email{690367242@qq.com}
\author[1,2]{\fnm{Xin} \sur{Li}}\email{782045777@qq.com}
\author[1,2]{\fnm{Yongfei} \sur{Lv}}\email{henulvyongfei@163.com}


\author[3]{\fnm{Chongsheng} \sur{Zhang}}\email{cszhang@henu.edu.cn}

\affil[1]{\orgdiv{School of Information Engineering}, \orgname{Henan Kaifeng College of Science Technology and Communication}, \orgaddress{ \city{Kaifeng}, \postcode{475000}, \state{Henan}, \country{China}}}

\affil[2]{\orgname{Henan Engineering Research Center for AI Technology of Converged Media}, \orgaddress{ \city{Kaifeng}, \postcode{475000}, \state{Henan}, \country{China}}}

\affil[3]{\orgdiv{School of Computer and
Information Engineering}, \orgname{Henan University}, \city{Kaifeng}, \postcode{475001}, \state{Henan}, \country{China}}



\abstract{
Biological learning systems acquire knowledge incrementally through broad, reusable representations established early and kept largely stable, while integrating new experience with minimal interference with previously learned knowledge. Replay-based methods approximate this by storing samples or features from previous tasks and mixing them into the current task. However, memory buffers grow with the number of tasks and may retain task-specific noise that reduces generalization. To address these limitations, we propose PatchAdaNet (Patch Adapter Network), an unsupervised class-incremental learning method for image classification that continually updates feature representations using autoencoders. Since images share structural features such as edges, corners, and shapes, we first divide each image into patches and train independent autoencoders on external data to reconstruct each patch. The layers of all patch encoders are aggregated into global frozen layers that generalize across tasks and remain fixed throughout learning. Learnable adapters then build global image representations from these features to reconstruct the entire image. To prevent feature drift without storing data, we introduce the Global Feature Centroid (GFC), a fixed-size running average of adapter outputs updated after each task to penalizes representation drift. Elastic Weight Consolidation (EWC) is applied to the learnable layers to preserve previously learned features by constraining important weights. Finally, a linear classifier is trained on the frozen latent representation and GFC for class-incremental evaluation. We evaluate PatchAdaNet in both CNN and Vision Transformer settings on CIFAR-10, CIFAR-100, miniImageNet and 5-Datasets, providing a comparative study of the two architectures under identical class-incremental settings. Code is available$^{1}$.}




\maketitle

\section{Introduction}\label{sec1}
When a neural network is trained sequentially, the parameter updates required to fit the current task conflict with the weight configurations that encode knowledge of previous tasks, a fundamental and largely unsolved challenge in deep learning known as catastrophic forgetting. This problem is especially severe in class-incremental learning (CIL), where the model must classify all classes seen so far without access to task identity at inference time. Existing continual learning methods address forgetting from several perspectives. Regularization methods constrain parameter updates by penalizing changes to weights identified as important for prior tasks \cite{kirkpatrick2017overcoming,zenke2017continual,aljundi2018memory}. For example, Elastic Weight Consolidation (EWC) \cite{kirkpatrick2017overcoming} measures weight importance through computing Fisher information and adds a quadratic penalty on parameter drift. Synaptic Intelligence \cite{zenke2017continual} estimates importance online during training, without requiring a separate importance estimation pass after each task. Architectural expansion methods allocate new, task-specific parameters for each task, either by instantiating an entirely new set of layers per task \cite{rusu2016progressive} or by selectively retraining and expanding only the neurons relevant to each new task \cite{yoon2018lifelong}. Replay-based methods store or generate examples from previous tasks and incorporate them with current-task training \cite{lopez2017gradient,shin2017continual}. Knowledge distillation methods use a frozen copy of the model from the immediately preceding task as the teacher, transferring its soft predictions to constrain the student trained on the current task \cite{li2017learning,wen2024class}.

{\renewcommand{\thefootnote}{}\footnotetext{1 \url{https://github.com/adjkamel/PatchAdaNet}}}

Despite this progress, several limitations remain across existing approaches: 1) regularization methods such as EWC operate purely in weight space, applying a single importance estimate uniformly across all layers. This merges task-specific and shared representations within the same parameters, making it difficult to isolate what should change and what should be preserved. 2) architectural expansion methods introduce parameter or memory costs that scale with the number of tasks. 3) replay-based methods require a growing memory buffer to retain samples or features from previous tasks, and the retained data often carries task-specific noise that reduces generalization to new tasks. 4) knowledge distillation methods depend on keeping a copy of the previous model as a teacher, which doubles the memory requirement, and the distilled features tend to become increasingly task-specific as the sequence grows, losing the general representations needed for class-incremental classification. 5) Supervised continual learning methods require the classifier to be restructured each time new classes arrive, binding representation learning directly to the evolving label space and limiting flexibility in open-ended scenarios.

To address each of the previous limitations, we propose PatchAdaNet (Patch Adapter Network). To tackle the entanglement of shared and task-specific representations, PatchAdaNet explicitly separates the two through a two-stage design. In the first stage, which is performed once before sequential training begins, each image is divided into non-overlapping patches and an independent autoencoder (PatchAE) is trained to reconstruct each patch, where each PatchAE is assigned to a specific patch position index and trained across all pretraining images at that index. The layer-wise features of the patch autoencoders are then aggregated to form global frozen layers that serve as a domain-agnostic feature generalizer across tasks and remain unchanged throughout the entire learning process, without neither weight updates nor regularization, making them immune to forgetting by construction. In the second stage, learnable adapter layers refine the shared frozen aggregated features to learn task-specific global image representations capturing only what is novel to the current task while preserving the shared structure already encoded in the frozen layers. The resulting task-specific representations are then decoded by a decoder to reconstruct the full input image. To address the limitations of replay, PatchAdaNet maintains no memory buffer. Instead of storing past samples or features, it maintains the Global Feature Centroid (GFC), a proposed compact fixed-size running average of adapter output layers updated after each task. The GFC serves as feature regularizer anchor to penalize representational drift away from the accumulated shared representation across tasks without retaining any data. EWC is applied exclusively to the adapter layers rather than to the frozen layers to keep the weight-space regularization target narrow and well-defined rather than spread across the full network, which reduces parameter interference. GFC  derive its regularization signal entirely from the model's own training process, without storing past samples or features, or relying on a separate teacher model. The representation is learned without any supervision, and a linear classifier is trained on the concatenation of the frozen bottleneck encoder output and the GFC features after each task, ensuring that feature learning remains independent of the expanding class label space. In addition, we provide a comparative analysis of class-incremental learning performance under the same framework using two different model architectures, CNN and Vision Transformer autoencoders.

The contributions of this work can be summarized in the following points:

\begin{itemize}

\item PatchAdaNet is proposed as a two-stage unsupervised continual learning framework that explicitly separates universal structural feature learning from task-specific adaptation. Patch-level autoencoders are pretrained once on external data and frozen permanently, providing a task-invariant feature base that eliminates forgetting in the shared representation.

\item Learnable adapter layers are introduced on top of the frozen features to learn task-specific representations, regularized by two complementary mechanisms: the GFC penalizes representational drift in feature space, while EWC constrains parameter drift in weight space. Both are derived exclusively from the model's own training process, without any external memory buffer or feature distillation.

\item PatchAdaNet is instantiated with both CNN and Vision Transformer architectures and evaluated on CIFAR-10, CIFAR-100, miniImageNet, and 5-Datasets under unsupervised class-incremental settings, with a comparative analysis of the two architectures. The experimental results demonstrate competitive performance against supervised and replay-based methods.

\end{itemize}

\begin{figure*}
\centering
\includegraphics[width=0.8\linewidth]{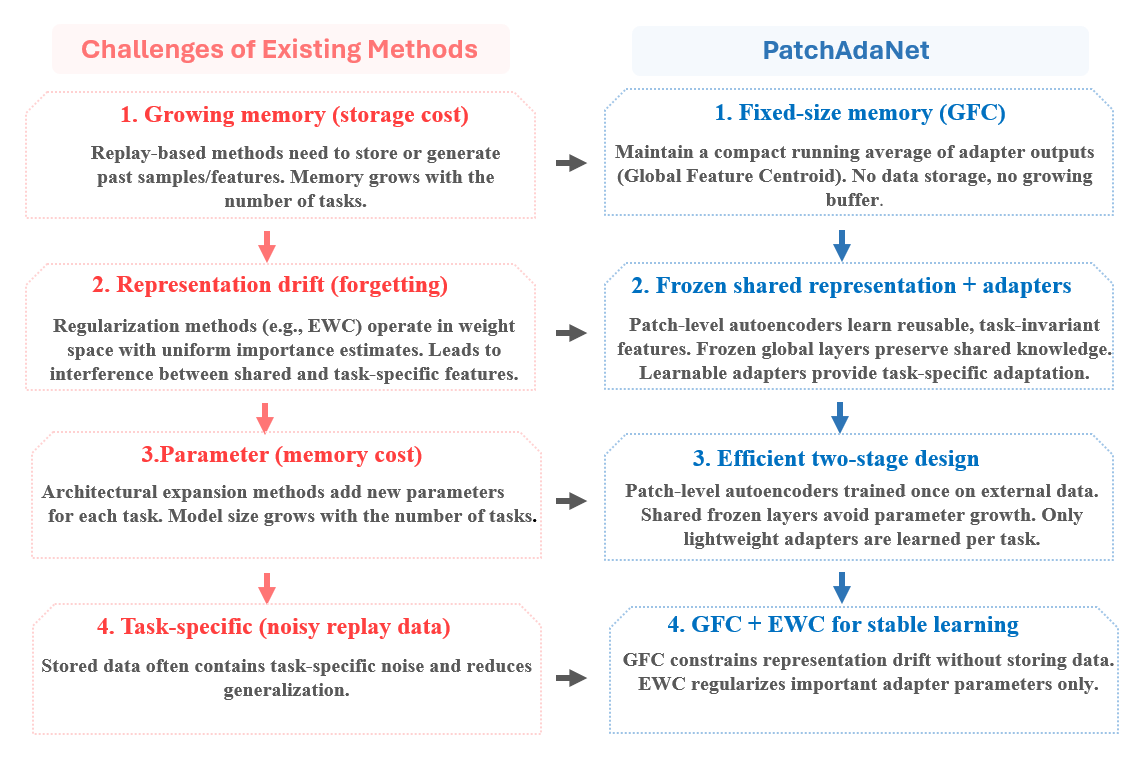}
\caption{Challenges of existing methods, and solutions of PatchAdaNet.}
\label{fig:learn} 
\end{figure*}

\section{Related Work}

Regularization-based approaches restrict how parameters are updated in order to preserve weights considered important for earlier tasks. Elastic Weight Consolidation (EWC) approximates parameter importance through Fisher information and penalizes deviation in critical weights by applying a quadratic penalty proportional to each weight's estimated importance \cite{kirkpatrick2017overcoming}. Synaptic Intelligence instead accumulates path-integrated importance scores  during training to track each parameter's contribution to the loss reduction as it updates \cite{zenke2017continual}. Uncertainty-aware regularization lowers the cost of per weight regularization by operating at the node level through estimated uncertainty, using a variational framework to scale regularization strength per unit rather than per parameter \cite{ahn2019uncertainty}, whereas Bayesian generative regularization \cite{chen2021overcoming} stabilizes continual updates by introducing probabilistic priors over the parameters, framing forgetting mitigation as posterior approximation across tasks. Replay-based approaches instead counter forgetting by revisiting stored or synthetically generated examples from earlier tasks. Strong experience replay  \cite{shin2017continual} distills knowledge from past experiences held in a memory buffer to maintain backward consistency, while additionally enforcing forward consistency by matching the previous model's outputs on current training data, combining buffer-based replay with a distillation objective.  Generative replay \cite{lopez2017gradient} achieved a similar effect by training a generative model to synthesize past-task samples rather than retaining raw data to avoid the need for an explicit memory buffer. Adversarial continual learning further incorporates adversarial training objectives to keep data distributions aligned across tasks, using a discriminator to encourage task-invariant feature representations \cite{ebrahimi2020adversarial}. Architectural strategies take a different route by expanding or restructuring the network itself to provide dedicated capacity for new tasks. Progressive neural networks grow the architecture as each new task is introduced, instantiating a new column of layers per task while freezing previous columns and using lateral connections to transfer knowledge \cite{rusu2016progressive}, while increasing depth strategy \cite{kozal2023increasing} instead append new layers over time to isolate knowledge specific to each task, expanding network depth rather than width as tasks accumulate.

Self-supervised contrastive learning techniques \cite{zbontar2021barlow}, \cite{chen2020simple} have also been applied to counter forgetting through continually learned representations. \citet{cha2021co2l} learn representations via contrastive learning and retain them through self-supervised distillation, using an asymmetric loss between current and past embeddings to prevent representation drift. \citet{fini2022self} demonstrate how a self-supervised loss can be repurposed as a distillation mechanism that aligns the current representation state with its past state, framing the distillation target as the output of the same self-supervised objective applied to a frozen prior model. \citet{wen2024provable} similarly combine contrastive loss with knowledge distillation, and further introduce an adaptive distillation scheme that assigns task-specific coefficients based on the similarity between tasks. \citet{cheng2023cucl} introduce a codebook for feature representation to strengthen discriminative power during continual learning, applying a contrastive loss across both the raw representations and the codebook features, using the codebook as a compact, quantized memory of representative features. Other recent work avoids relying on stored past information altogether. \citet{cheng2024must} proposed Alternate Task Discrimination, an information-free approach built on a self-supervised pretext task, which improves resistance to forgetting by training the model to identify which data stream a sample originates from via contrastive learning on augmented data, without access to task identity or stored exemplars at training time.

Distinct from these directions, our approach is replay-free and depends on neither contrastive learning, pretrained models, nor distillation. It relies solely on independently trained, regularized representations for image reconstruction.

\begin{figure*}
\centering
\includegraphics[width=1\linewidth]{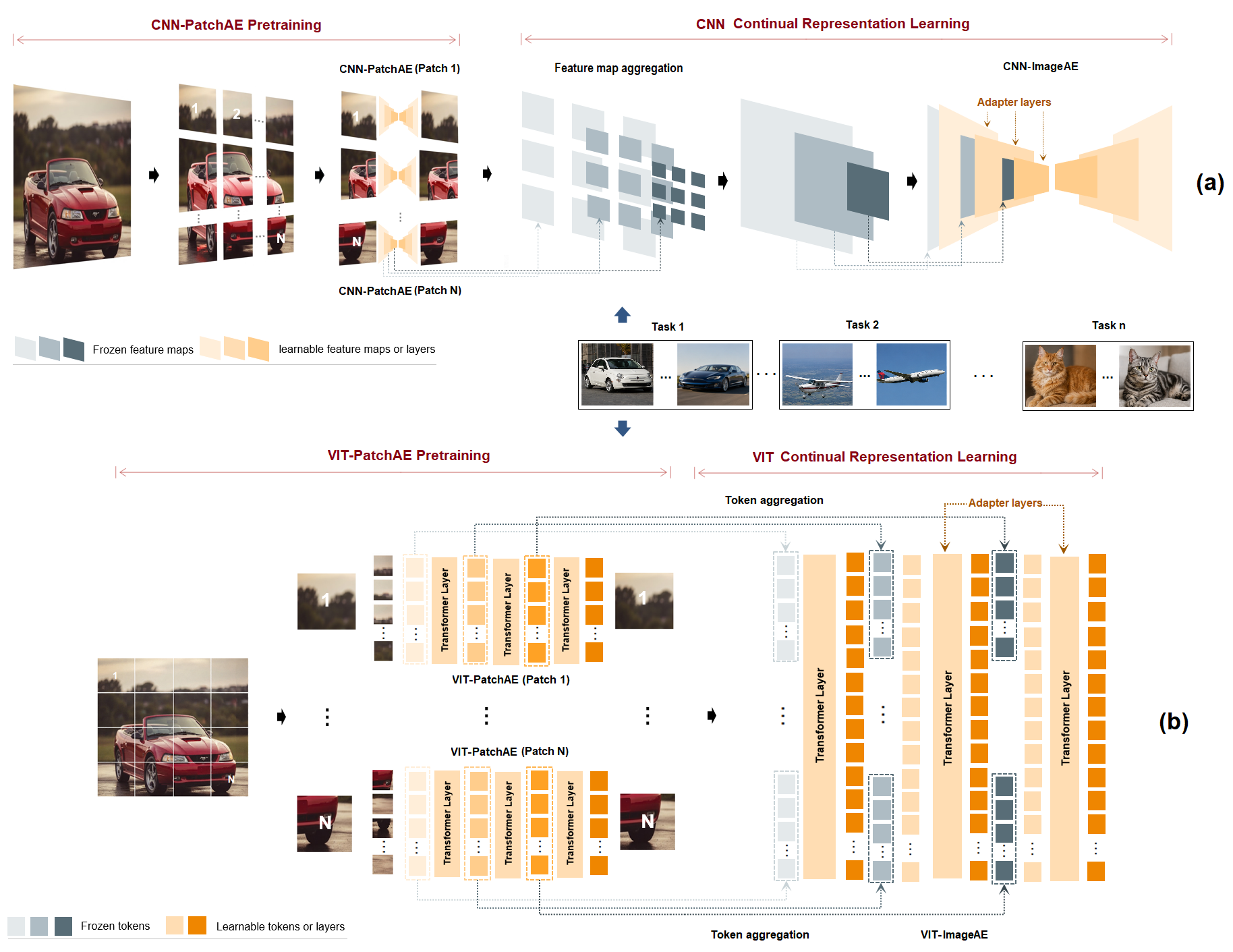}
\caption{Visual illustration of the proposed PatchAdaNet. \textbf{(a)} CNN-based framework: frozen patch-level features are pretrained using a patch-level CNN autoencoder (CNN-PatchAE); continual learning then refines tiled frozen feature maps through adapter layers via image reconstruction (CNN-ImageAE). \textbf{(b)} ViT-based framework: frozen patch-level features are pretrained using a patch-level ViT autoencoder (ViT-PatchAE). In continual learning phase, frozen tokens are concatenated and combined with the previous depth's adapted tokens,  then refined through adapter Transformer blocks via image reconstruction (ViT-ImageAE). }
\label{fig:frame} 
\end{figure*}

\section{Method}
\subsection{Overview} 

Figure \ref{fig:frame} shows a visual illustration of the proposed idea of PatchAdaNet. We investigate two approaches: a CNN approach and a Vision Transformer (ViT) approach, which both operate through the same processing steps but with different internal structures. The fundamental idea is to learn small independent representations once, that are shared across sequential tasks through individual patch-level pretraining. These low- to mid-level structural features, such as edges, corners, and textures, are common to any natural image and therefore generalizable across all tasks. After pretraining, the patch-level features are aggregated to form a global, image-level, frozen shared patch-level feature extractor across tasks. This frozen feature extractor is followed by adapter layers that continually learn task-specific image-level features. We investigate two approaches: CNN-based approach and Vision Transformer (ViT) approach.

In the case of the CNN approach, the frozen patch-level feature maps produced by CNN-PatchAE are spatially tiled and aggregated to form the global frozen feature extractor at different layer depths, then a global image-level CNN-ImageAE autoencoder trained continually across tasks. Learnable adapter layers are placed after each frozen layer depth to refine the aggregated feature maps for the current task. In the case of the ViT approach, the frozen patch-level token sequences produced by ViT-PatchAE are concatenated to form the global frozen feature extractor, which serves as the input to ViT-ImageAE. At each depth, the frozen token sequence is combined with the previous depth's adapted tokens and refined through adapter Transformer blocks. 

During continual learning, EWC is applied exclusively to the adapter and reconstruction layers of CNN-ImageAE and ViT-ImageAE, leaving the frozen CNN-PatchAE and ViT-PatchAE layers entirely unaffected. To further regularize the adapter outputs across tasks without storing any data, the introduced Global Feature Centroid (GFC) keeps a running average of the adapter at each depth that is updated after every task. During training, this average acts as a drift penalty that pulls the current adapter outputs toward it to keep the representation stable across tasks. After each task, a linear classifier is trained on the frozen bottleneck representation of CNN-ImageAE or ViT-ImageAE with the GFC features for class-incremental evaluation. Figure~\ref{fig:learn} illustrates the training process in detail, showing the three loss terms applied during continual learning: the image reconstruction loss, the GFC drift loss, and the EWC penalty.

\subsection{Patch partitioning}
As a prior step of the sequential learning process across tasks, we perform pretraining on small dataset. This pretraining is conducted once, then the pretrained features are used for all tasks, and for all datasets to capture low-level to mid-level visual structures.

\begin{figure*}
\centering
\includegraphics[width=1\linewidth]{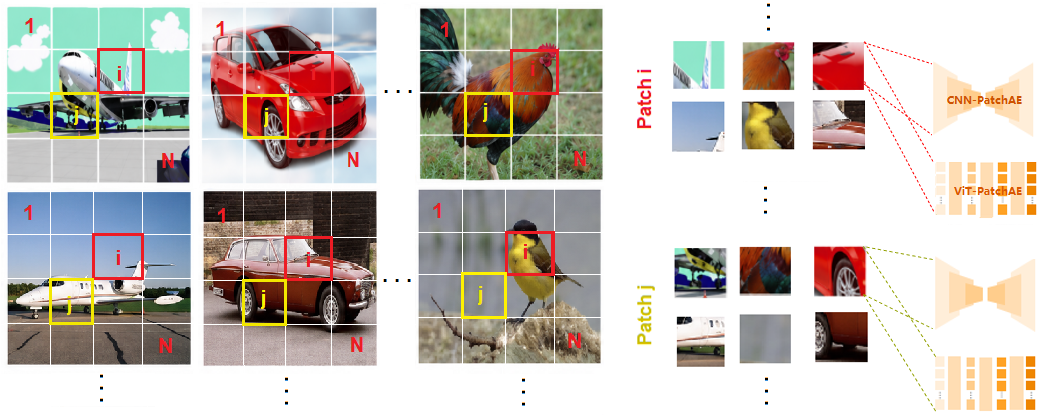}
\caption{Illustration of patch partitioning and training a dedicated autoencoder per patch index using CNN-PatchAE or ViT-PatchAE.}
\label{fig:patch} 
\end{figure*}

Let $T_1,T_2, \dots, T_k$ be a sequence of tasks, where the data of each task is $D_k = \left\{ \left( x_i^k,\, y_i^k \right) \right\}_{i=1}^{n_k}$. The learned $r$ representations  of each task are $Z_{1k},Z_{2k},\dots,Z_{rk}$ using $r$ autoencoders to reconstruct the input. Given an input image $x \in \mathbb{R}^{C \times H \times W}$, we partition non-overlapping patches of equal size. Each patch $p_i \in \mathbb{R}^{C \times h_p \times w_p}$ is obtained by dividing $x$ into spatially contiguous regions of size $h_p \times w_p$, where $h_p = H/R$ and $w_p = W/R$ ($R\times R$ is the total number of patches), such that the patches together cover the entire image without overlap. Each patch occupies a fixed spatial position in the image and is assigned its own dedicated autoencoder, trained on patches from that specific spatial position across all images in the pretraining dataset. Figure \ref{fig:patch} provides a visual illustration of the patch partitioning process.

\subsubsection{CNN Patch Auto-Encoder (CNN-PatchAE)}

Each patch $p_i$ is independently encoded by a dedicated convolutional encoder 
$\mathrm{Enc}_i$, which maps the patch to a hierarchy of feature representations 
across $D$ encoding depths. At each depth $d$, the encoder produces a feature map 
by applying a convolutional layer followed by a non-linear activation:

\begin{equation}
    z_i^d = \sigma\bigl(W_d \ast z_i^{d-1} ), 
    \quad d = 1, \ldots, D, 
    \quad z_i^0 = p_i
\end{equation}

where $W_d$ is the learnable convolutional kernel at depth $d$, $k$ is the kernel size, $\ast$ denotes the convolution operator, and $\sigma(\cdot)$ denotes the ReLU activation function. The number of feature channels  at depth $d$ is $C_d \in \{F, 2F, 4F, \dots\}$, where $C_d = 2^{d-1} F$ and $F \in \mathbb{Z}^+$ is a base channel width, reflecting a progressively richer representation with depth. A symmetric decoder $\mathrm{Dec}_i$ reconstructs the patch from the bottleneck representation $z_i^D \in \mathbb{R}^{C_DF \times h_D \times w_D}$ through a sequence 
of transposed convolutions. Each $\mathrm{PatchAE}_i = (\mathrm{Enc}_i, \mathrm{Dec}_i)$ is trained independently on external pretraining data by minimizing the pixel-level reconstruction loss over all $P$ patches:

\begin{equation}
    \mathcal{L}_{\mathrm{patch}} = \frac{1}{P} \sum_{i=1}^{P} 
    \left\| \mathrm{Dec}_i\!\left(z_i^D\right) - p_i \right\|_F^2
\end{equation}

Since each encoder $\mathrm{Enc}_i$ reconstructs only the local spatial region at position $i$, without access to the surrounding image context, it is implicitly constrained to capture low to mid-level structural features such as edges, corners, and local textures that are common to natural images across tasks and domains. Upon completion of pretraining, the encoder parameters $\theta_i^{\mathrm{enc}} = \{W_d,\, b_d\}_{d=1}^{D}$ of all $P$ patch encoders are frozen and remain unchanged throughout the continual learning process to ensure that the patch-level representations $\{z_i^d\}_{d=1}^{D}$ remain fixed and task-invariant throughout the entire continual learning process across all tasks.

\subsubsection{ViT Patch Autoencoder (ViT-PatchAE)}

In the Transformer variant, the same patch partitioning strategy is applied, however, instead of convolutional feature maps, each patch is represented as a sequence of tokens processed by an independent Vision Transformer autoencoder. Given the $i$-th patch $p_i \in \mathbb{R}^{C \times h_p \times w_p}$, we further divide it into $N_i = (h_p / s) \times (w_p / s)$ non-overlapping sub-patches of size $s \times s$ 
pixels. Each sub-patch is linearly projected into a $d_s$ dimensional embedding space, and a learnable positional embedding 
$\mathbf{E}^{\mathrm{pos}} \in \mathbb{R}^{N_i \times d_s}$ is added to encode spatial order within the patch:

\begin{equation}
    \mathbf{T}_i^0 = \mathrm{Proj}(\mathrm{tokenize}(p_i)) + 
    \mathbf{E}^{\mathrm{pos}} \in \mathbb{R}^{N_i \times d_s}
\end{equation}

where $\mathrm{Proj}: \mathbb{R}^{c \times s \times s} \to \mathbb{R}^{d_s}$ with $c$ represents image channels, is a learned linear projection applied to each flattened sub-token. The token sequence $\mathbf{T}_i^0$ is then passed through $D$ successive Transformer encoder blocks, each consisting of multi-head self-attention and a feed-forward network. As in the CNN case, each block corresponds to one encoding depth $d = 1, \ldots, D$, so that the output of the $d$-th block represents 
the patch's token-level representation at depth $d$:

\begin{equation}
    \mathbf{T}_i^d = \mathrm{TransformerBlock}_d\!\left(\mathbf{T}_i^{d-1}\right) 
    \in \mathbb{R}^{N_i \times d_s}
\end{equation}

The output token sequence at each block $d$ constitutes a depth-level representation of the patch, analogous to the layer-wise feature maps in the CNN case. A symmetric Transformer decoder reconstructs the patch from the deepest token sequence 
$\mathbf{T}_i^D$ by projecting each token back to pixel space via a learned projection $\mathrm{Dec\_proj}: \mathbb{R}^{d_s} \to \mathbb{R}^{c \times s \times s}$.

\begin{equation}
    \mathcal{L}_{\mathrm{patch}} = \frac{1}{P} \sum_{i=1}^{P} 
    \left\| \mathrm{Dec}\!\left(\mathbf{T}_i^D\right) - p_i \right\|_F^2
\end{equation}

Since each ViT-PatchAE operates solely on the tokens of its assigned spatial patch, the self-attention mechanism is restricted to local intra-patch token interactions. As with the CNN variant, this allows the encoder to encode low to mid-level local structural features. Similarly, upon completion of pretraining, the encoder parameters $\theta_i^{\mathrm{enc}} = \{\mathbf{W}_d, \mathbf{E}^{\mathrm{pos}}\}_{d=1}^{D}$ of all $P$ patch encoders are frozen to ensure that the token-level structural representations $\{\mathbf{T}_i^b\}_{b=1}^{B}$ remain fixed and task-invariant throughout the continual learning process.

\begin{figure*}
\centering
\includegraphics[width=1\linewidth]{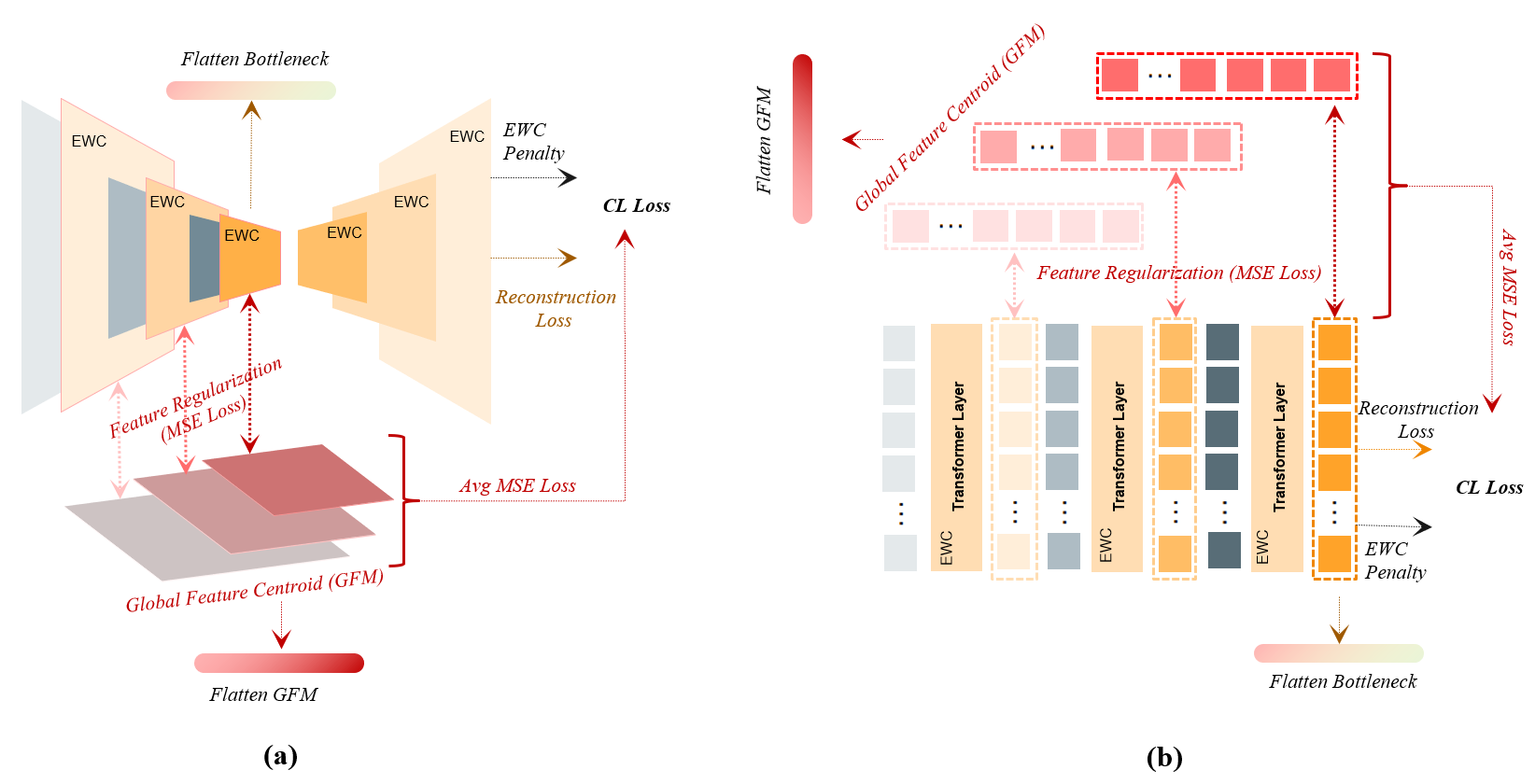}
\caption{Detailed illustration of PatchAdaNet training process in CNN settings (a), and in ViT settings (b). A linear classifier is trained on top of the resulting Flatten GFM and Flatten Bottleneck features for class incremental learning evaluation.}
\label{fig:learn} 
\end{figure*}

\subsubsection{CNN Image Auto-Encoder (CNN-ImageAE)}

Given the frozen encoder output of CNN-PatchAE at depth $d$ for patch $i$,  denoted $z_i^d$, a padded variant of the frozen convolution is first applied to preserve spatial dimensions:

\begin{equation}
    f_i^d = \sigma\bigl(\tilde{W}_d \ast z_i^{d}), \quad f_i^1 = p_i
\end{equation}

where $\tilde{W}_d$ is the frozen kernel corresponding CNN-PatchAE encoder, with padding added to recover the patch's original spatial resolution. The frozen feature maps $\{f_i^d\}_{i=1}^{P}$ at depth $d$ are spatially tiled according to each patch's grid position to reconstruct a full image size frozen feature map:

\begin{equation}
    \mathbf{G}_{\mathrm{frozen}}^d = \mathrm{tile}\bigl(\{f_i^d\}_{i=1}^{P}\bigr) 
    \in \mathbb{R}^{C_d \times H_d \times W_d}
\end{equation}

On top of $\mathbf{G}_{\mathrm{frozen}}^d$, a learnable adapter module $\mathcal{A}_d$ is applied to produce the task-specific adapted representation at depth $d$. Each adapter $\mathcal{A}_d$ consists of $L$ sequential convolutional layers of fixed channel width $C_d$, each followed by a ReLU activation:

\begin{multline}
    \mathbf{G}_{\mathrm{adapt}}^d = \mathcal{A}_d\bigl(\mathbf{G}_{\mathrm{frozen}}^d\bigr) \\
    = \sigma\Bigl(W_d^{(L)} \ast \cdots \ast
      \sigma\bigl(W_d^{(1)} \ast \mathbf{G}_{\mathrm{frozen}}^d \bigr) \dots\Bigr)
\end{multline}

where $\{W_d^{(\ell)}, b_d^{(\ell)}\}_{\ell=1}^{L}$ are the learnable parameters of adapter $\mathcal{A}_d$, and $L$ denotes the number of repeated adapter layers at depth $d$. Unlike the frozen feature maps $\mathbf{G}_{\mathrm{frozen}}^d$, the adapter parameters $\theta_d^{\mathrm{adapt}} = \{W_d^{(\ell)}, b_d^{(\ell)}\}_{\ell=1}^{L}$ are updated continually across all $K$ tasks. The depth-wise sequence of frozen-feature  followed by adapter refinement together constitutes the image encoder, denoted $\mathrm{Enc}_{\mathrm{img}}$.

The deepest adapted representation $\mathbf{G}_{\mathrm{adapt}}^{D}$ serves as the global bottleneck and is passed to a decoder $\mathrm{Dec}_{\mathrm{img}}$, composed of transposed convolutions, to reconstruct the full input image:

\begin{equation}
    \hat{x} = \mathrm{Dec}_{\mathrm{img}}\bigl(\mathbf{G}_{\mathrm{adapt}}^{D}\bigr)
\end{equation}

The global reconstruction loss for task $k$ is given by:

\begin{equation}
    \mathcal{L}_{\mathrm{recon}}^{(k)} = \left\| \hat{x} - x \right\|_F^2
\end{equation}

Because only the adapter parameters $\theta_d^{\mathrm{adapt}}$ receive gradient updates while the frozen feature maps $\mathbf{G}_{\mathrm{frozen}}^d$ remain constant across all tasks, catastrophic forgetting in CNN-AdaAE is confined entirely to the adapter layers, with the frozen CNN-PatchAE representations acting as a stable, task-invariant feature substrate throughout continual learning.

\subsubsection{ViT Image Auto-Encoder (ViT-ImageAE)}

In the second stage of the ViT-based approach, the frozen patch-level token sequence representations produced by ViT-PatchAE at each depth $d$ for each patch $p_i$, denoted $\mathbf{S}_i^d \in \mathbb{R}^{N_i \times \dim}$, are concatenated across all $P$ patches to form a global frozen sequence  $\mathbf{G}_{\mathrm{frozen}}^d$, and a learnable depth-specific positional embedding $\mathbf{E}^{\mathrm{pos}}_d \in \mathbb{R}^{N_{\mathrm{total}} \times \dim}$ is added:

\begin{equation}
\begin{aligned}
    \mathbf{G}_{\mathrm{frozen}}^d = \mathrm{cat}\bigl(\{\mathbf{S}_i^d\}_{i=1}^{P},\; 
    \mathrm{dim}{=}1\bigr) + \mathbf{E}^{\mathrm{pos}}_d
\end{aligned}
\end{equation}

where $N_{\mathrm{total}} = P \times N_i$ is the total number of tokens across all patches at depth $d$. At each depth, the token sequence $\mathbf{G}_{\mathrm{frozen}}^d$ is passed through a trainable Transformer adapter block $\mathrm{AdapterBlock}_d$ consisting of multi-head self-attention and a feed-forward network, which produces a learned token sequence that is then added to the frozen tokens at the next depth.

\begin{equation}
    \mathbf{G}_{\mathrm{adapt}}^{d+1} = \mathrm{AdapterBlock}_d\bigl(\mathbf{G}_{\mathrm{frozen}}^d + 
    \mathbf{G}_{\mathrm{adapt}}^{d-1}\bigr)
\end{equation}

At depth $d = 1$, no previous adapter output exists, so the frozen tokens at depth 1 are passed directly into the fusion block without addition:

\begin{equation}
    \mathbf{G}_{\mathrm{adapt}}^{1} = \mathrm{AdapterBlock}_1\bigl(\mathbf{G}_{\mathrm{frozen}}^1\bigr)
\end{equation}

Each  $\mathrm{AdapterBlock}_d$ can be followed by Transformer blocks with the same structure to increase feature representations before processing the next $\mathbf{G}_{\mathrm{frozen}}^{d}$ sequence can be followed by Transformer blocks with the same structure to increase feature representation. Both the fusion blocks. All  $\mathrm{AdapterBlock}_d$ are fully trainable and are updated continually across all $K$ tasks, with parameters $\theta_d^{\mathrm{fuse}}$ and $\theta_d^{\mathrm{adapt}} = \{\mathbf{W}_d^{(\ell)}\}_{\ell=1}^{L}$, respectively. The only parameters held fixed throughout continual learning are $\mathbf{G}_{\mathrm{frozen}}^{d}$. 

The deepest adapter output $\mathbf{G}_{\mathrm{adapt}}^{D}$ serves as the global bottleneck representation. It is normalized and projected through a sequence of Transformer decoders $\mathrm{DecBlock}$, followed by a final projection $\mathrm{dec\_proj}: \mathbb{R}^{\dim} \to \mathbb{R}^{c \times s\times s}$ that maps each token back to pixel space. The resulting token grid is reassembled into 
the full image via an inverse patchify operation $\mathrm{unpatchify}(\cdot)$:

\begin{equation}
\begin{aligned}
    \hat{x} &= \mathrm{unpatchify}\Bigl(\mathrm{dec\_proj}\bigl( \mathrm{DecBlock}(\cdots \mathrm{DecBlock}(\mathbf{G}_{\mathrm{adapt}}^{D})\cdots)\bigr)\Bigr)
\end{aligned}
\end{equation}

The global reconstruction loss for task $k$ is given by:

\begin{equation}
    \mathcal{L}_{\mathrm{recon}}^{(k)} = \left\| \hat{x} - x \right\|_F^2
\end{equation}

\subsubsection{Global Feature Centroid (GFC)}

To regularize the adapter outputs across the task sequence without storing any samples, we introduce the \textbf{Global Feature Centroid (GFC)}: a set of $D$ running-average tensors $\mu_1, \mu_2, \ldots, \mu_D$, one per encoding depth. For the CNN backbone, $\mu_d \in \mathbb{R}^{1 \times C_d \times H_d \times W_d}$ matches the spatial shape of the adapted feature map $\mathbf{G}_{\mathrm{adapt}}^d$; for the ViT backbone, $\mu_d \in \mathbb{R}^{1 \times N_{\mathrm{total}} \times \dim}$ matches the shape of the adapted token sequence $\mathbf{G}_{\mathrm{adapt}}^d$. In both cases, the GFC is constructed not from the full training set, but exclusively from the last mini-batch processed during the final training epoch of each task. A design choice that limits the construction cost to a single forward pass per task rather than a full pass over the dataset.

After the final training epoch on task $T_1$ completes, the GFC is initialized using only the last mini-batch seen during that epoch. The frozen representation $\mathbf{G}_{\mathrm{frozen}}^d$ computed for that single batch frozen feature maps for the CNN backbone and frozen token sequences for the ViT backbone is averaged over the batch dimension:
\begin{equation}
    \mu_d^{(1)} = \mathrm{mean}_D\bigl(\mathbf{G}_{\mathrm{frozen}}^d)
\end{equation}

After the final training epoch on task $T_k$ ($k>1$) completes, $\mu_d$ is updated using only the adapted representation $\mathbf{G}_{\mathrm{adapt}}^d$ from the last mini-batch of that epoch, via Exponential Moving Average (EMA) with momentum $m = 0.9$:

\begin{equation}
\begin{aligned}
    \mu_d^{(k)} &\leftarrow m \cdot \mu_d^{(k-1)} + (1-m) \cdot \mathrm{mean}_B\bigl(\mathbf{G}_{\mathrm{adapt}}^d\bigr)
\end{aligned}
\end{equation}

As with initialization, this update depends solely on the single last batch of task $T_k$, not an average over all batches encountered during training on that task, and the update rule is identical for both backbones. The update is performed without gradient tracking, so $\mu_d$ accumulates a smoothed cross-task average, with a fixed memory footprint of $D$ tensors regardless of the number of tasks $K$.

During training on any task $T_k$ with $k>1$, $\mu_d$ is constructed from a last-batch snapshot of the previous task, serves as a fixed regularization target for $\mathbf{G}_{\mathrm{adapt}}^d$ at every training iteration of the current task. At each forward pass, $\mu_d$ is detached from the computation graph, and a mean-squared-error penalty loss is computed between $\mathbf{G}_{\mathrm{adapt}}^d$ and 
$\mu_d$ at every depth:
\begin{equation}
    \mathcal{L}_{\mathrm{GFC}}^d = \left\| \mathbf{G}_{\mathrm{adapt}}^d - 
    \mathrm{sg}(\mu_d) \right\|_F^2
\end{equation}
\begin{equation}
    \mathcal{L}_{\mathrm{GFC}} = \frac{1}{D} \sum_{d=1}^{D} \mathcal{L}_{\mathrm{GFC}}^d
\end{equation}

\subsection{Total loss function}
\label{sec:loss_summary}

During continual learning across tasks using CNN-ImageAE or ViT-ImageAE, the model is optimized using a total loss that combines image reconstruction, feature-space drift regularization (GFC), and weight regularization via Elastic Weight Consolidation (EWC), which penalizes changes to weights that were important for previous tasks.

The adapter layers are  constrained in weight space relative to their values after the previous task, weighted by accumulated Fisher information:
\begin{equation}
    \mathcal{L}_{\mathrm{EWC}}^{(k)} = \sum_n \mathcal{F}_n \left(\theta_n - 
    \theta_n^{*}\right)^2
\end{equation}
For task $T_1$, $\mathcal{L}_{\mathrm{EWC}}^{(1)} = 0$, since no prior task exists 
to constrain against.

The overall loss minimized during training on task $T_k$ is the weighted sum of the active terms:
\begin{equation}
    \mathcal{L}_{\mathrm{total}}^{(k)} = \mathcal{L}_{\mathrm{recon}}^{(k)} 
    + \alpha \cdot \mathcal{L}_{\mathrm{GFC}}^{(k)} 
    + \lambda \cdot \mathcal{L}_{\mathrm{EWC}}^{(k)}
\end{equation}
where $\alpha$ and $\lambda$ weight the penalty imposed during training for GFC and EWC, respectively.

\section{Experiments}

\subsection{Datasets and evaluation metrics}

We evaluate and compare our method mainly on conventional task splits of 5-Split CIFAR-10 with 2 classes per task, 20-Split CIFAR-100 and 20-Split miniImageNet, each with 5 classes per task. We report classification accuracy under the task-incremental learning (TIL) or class-incremental learning (CIL) settings. Evaluation is conducted for PatchAdaNet in both CNN-based and ViT-based variants using a linear classifier trained on the frozen bottleneck features of the CNN-ImageAE or ViT-ImageAE encoders. To assess robustness, we evaluate the sensitivity of both variants to two sources of randomness, the random seed, by repeating each run with different seeds at a fixed class order (15 seeds on CIFAR-10 and CIFAR-100, 5 on miniImageNet), and the class order, by repeating each run with different random class-to-task permutations at a fixed seed (5 permutations per setting). We further examine sensitivity to data domain order on the 5-Datasets sequence: : MNIST, SVHN, notMNIST, Fashion-MNIST, and CIFAR-10.

\subsection{Implementation details}

\label{sec:implementation}
The patch-level autoencoders (CNN-PatchAE and ViT-PatchAE) are pretrained once on an external dataset (TinyImageNet), and kept entirely disjoint from all continual learning evaluation datasets. This design choice tests whether the low to mid-level structural features learned at the patch level are generic enough to transfer across datasets that the patch encoders never directly observed during pretraining.

In both CNN and ViT settings, images are resized to $64 \times 64$ pixels for all datasets. The image is divided into $P = 16$ patches (default $4\times4$ grid), each trained with an independent CNN-PatchAE or ViT-PatchAE branch for $20$ epochs with learning rate $\mathrm{lr} = 10^{-4}$ using Adam and a ReduceLROnPlateau scheduler (decay factor $0.8$, patience $5$). For each task $T_k$, the global model (CNN-ImageAE or ViT-ImageAE) is trained for $\mathrm{epochs} = 30$ epochs with Adam at learning rate $\mathrm{lr} = 10^{-4}$, batch size $128$, and the same  plateau scheduler configuration. Each depth uses $L = 3$ adapter layers. The GFC loss is weighted by $\alpha = 0.5$, and the EWC penalty is weighted by $\lambda = 5000$ in the CNN setting and $\lambda = 10$ in the ViT setting. GFC tensors are saved per task and reloaded as initialization for the next. 

In the CNN setting, the feature dimension is set to $F = 64$, yielding channel widths $\{64, 128, 256\}$ across the three encoding depths ($D=3$). When training CNN-ImageAE, only the tiled frozen CNN-PatchAE feature maps are excluded from the optimizer during global model training. In the ViT setting, each patch is tokenized with patch size $s = 4$ pixels and embedded into a $\mathrm{dim} = 128$ dimensional space, processed by $D = 3$ Transformer encoder blocks (depth), with $4$ attention heads. Unlike the CNN backbone, the encoder blocks and positional 
embeddings inside ViT-ImageAE are fully trainable and updated by the same optimizer as the adapter blocks, and only the external ViT-PatchAE is held frozen.

After CNN-ImageAE or ViT-ImageAE training completes for each task $T_k$, a linear classifier is trained on frozen features formed by concatenating the flattened bottleneck encoder output and the flattened deepest GFC centroid, $\mathbf{h} = \mathrm{cat}\!\left[\mathrm{flatten}\!\left(\mathbf{G}_{\mathrm{adapt}}^{D}\right),\; \mathrm{flatten}\!\left(\mathrm{sg}(\mu_{D})\right)\right]$,
where a single lineare layer $g_\phi: \mathbb{R}^h \to \mathbb{R}^{|\mathcal{C}|}$ is trained with cross-entropy loss using Adam ($\mathrm{lr} = 10^{-4}$, ReduceLROnPlateau scheduler, $150$ epochs). $|\mathcal{C}|$ represents the number of classes seen 
up to task $T_k$ in the case of CIL evaluation, and the number of classes of task $T_k$ only in the case of TIL evaluation.

\begin{table}[htbp]
\centering
\caption{%
  Class-Incremental Learning (CIL) results on CIFAR-10 (5~splits)
  and CIFAR-100 (20~splits) averaged over 15 runs with different random seeds.
  Memory: exemplar replay buffer (100 samples, all methods).
  \texttimes : replay-free.
}
\label{tab:supervised_cifar}
\renewcommand{\arraystretch}{1.2}
\begin{tabular}{@{}llccc@{}}
\toprule
\textbf{Method} & & \textbf{Memory} & \textbf{CIFAR-10} & \textbf{CIFAR-100} \\
\midrule
AGEM~\cite{chaudhryefficient}        && \checkmark & $17.7 \pm 0.3$ & $5.8 \pm 0.1$  \\
GSS~\cite{aljundi2019gradient}       && \checkmark & $18.4 \pm 0.2$ & $8.1 \pm 0.2$  \\
MIR~\cite{aljundi2019online}         && \checkmark & $20.7 \pm 0.7$ & $9.7 \pm 0.3$  \\
GDumb~\cite{prabhu2020gdumb}         && \checkmark & $23.3 \pm 1.3$ & $8.2 \pm 0.2$  \\
ASER~\cite{shim2021online}           && \checkmark & $20.0 \pm 1.0$ & $11.0 \pm 0.3$ \\
SCR~\cite{mai2021supervised}         && \checkmark & $40.2 \pm 1.3$ & $19.3 \pm 0.6$ \\
CoPE~\cite{de2021continual}          && \checkmark & $33.5 \pm 3.2$ & $11.6 \pm 0.7$ \\
DVC~\cite{gu2022not}                 && \checkmark & $35.2 \pm 1.7$ & $15.4 \pm 0.7$ \\
OCM~\cite{guo2022online}             && \checkmark & $47.5 \pm 1.7$ & $19.7 \pm 0.5$ \\
OnPro~\cite{wei2023online}  && \checkmark & $57.8 \pm 1.1$ & $22.7 \pm 0.7$ \\
Ours CNN  && \texttimes & \textbf{65.9 ± 0.7} & 22.3 ± 4.1 \\
Ours ViT  && \texttimes & \textbf{64.5 ± 0.4} & \textbf{32.0 ± 1.1} \\
\bottomrule
\end{tabular}
\label{tab:cifar} 
\end{table}

\subsection{Comparison}

\label{sec:comparison}

Table~\ref{tab:cifar} compares PatchAdaNet against supervised class-incremental learning methods on CIFAR-10 (5 splits) and CIFAR-100 (20 splits), using results from OnPro~\cite{wei2023online}. For a fair comparison, following OnPro, we report the average of 15 runs across random seeds. All compared methods rely on an exemplar replay buffer of 100 stored samples, whereas both PatchAdaNet variants are entirely replay-free. On CIFAR-10, PatchAdaNet achieves $65.9\%$ and $64.5\%$ accuracy in the CNN and ViT settings, respectively, clearly outperforming the OnPro replay-based baseline ($57.8 \pm 1.1\%$). On CIFAR-100, CNN attains $22.3\%$, on par with OnPro's $22.7 \pm 0.7\%$, while ViT reaches $32.0\%$, above the best replay-based result despite using no stored exemplars. These results indicate that the frozen patch-level feature foundation with adapters, combined with GFC and EWC regularization, can match or exceed replay-based methods without storing any past features or exemplars.

Table~\ref{tab:mini} reports task-incremental learning (TIL) results on miniImageNet under a 20-split protocol. Prior results are obtained from KCL~\cite{derakhshani2021kernel}. Following their evaluation settings, we report the average of 5 runs across random seeds. Both PatchAdaNet variants are replay-free and improve over the replay-free baselines Naive-SGD ($36.1 \pm 1.3\%$) and EWC alone ($34.8 \pm 2.3\%$). The CNN variant reaches $51.6\%$ and the ViT variant reaches $57.0\%$, the latter is better than the replay-based method KCL~\cite{derakhshani2021kernel} ($53.3 \pm 0.6\%$).

Table~\ref{tab:unsup} restricts the comparison to unsupervised class-incremental learning methods. On CIFAR-10, the CNN variant reaches the highest accuracy among the listed methods at $65.9\%$, outperforming the replay-based Co$^{2}$L~\cite{cha2021co2l} at $65.6 \pm 1.4\%$ and the distillation-based CaSSLe variants~\cite{fini2022self}, which reach $58.3\%$ and $62.7\%$ with the SimCLR and Barlow Twins backbones by relying on distilling from a frozen teacher model. The ViT variant reaches $64.5\%$, above both CaSSLe variants and slightly below Co$^{2}$L's replay-based result. On CIFAR-100, the ViT variant reaches the highest accuracy at $32.0\%$, above CLA (SimSiam)~\cite{pmlr-v330-cignoni26a} ($24.4 \pm 1.3\%$), which relies on a stored feature buffer and distilled features from the previous tasks.


 \begin{table}[htbp]
\centering
\caption{%
  Task-Incremental Learning (TIL) results on miniImageNet (20~splits) averaged over 5 runs with different random seeds. Memory: exemplar replay buffer. \texttimes: replay-free.
}
\label{tab:supervised_mini}
\renewcommand{\arraystretch}{1.2}
\begin{tabular}{@{}llcc@{}}
\toprule
\textbf{Method} & & \textbf{Memory} & \textbf{miniImageNet} \\
\midrule
Naive-SGD~\cite{mirzadeh2020understanding} && \texttimes   & $36.1 \pm 1.3$ \\
EWC~\cite{kirkpatrick2017overcoming}        && \texttimes   & $34.8 \pm 2.3$ \\
KCL~\cite{derakhshani2021kernel}   && \checkmark & $53.3 \pm 0.6$\\
Ours (CNN)  && \texttimes   & 51.6 ± 0.6 \\
Ours (ViT)      && \texttimes   & \textbf{57.0 ± 0.3} \\
\bottomrule
\end{tabular}
\label{tab:mini} 
\end{table}
 
\begin{table}[htbp]
\centering
\caption{%
  Unsupervised Class-Incremental Learning (CIL) results.
  M: buffer of features (20 samples) averaged over 15 runs with different random seeds.
  $\times$: replay-free. Dis: distillation from the previous tasks.
}
\label{tab:unsupervised_cifar}
\renewcommand{\arraystretch}{1.2}
\begin{tabular}{@{}llcc@{}}
\toprule
\textbf{Method} & & \textbf{Reply} & \textbf{Accuracy} \\
 
\midrule
\multicolumn{4}{@{}l}{\textbf{CIFAR-10}} \\
\midrule
Co$^{2}$L~\cite{cha2021co2l}                    && M & $65.6 \pm 1.4$ \\
CaSSLe (SimCLR)~\cite{fini2022self}&& Dis & $58.3$ \\
CaSSLe (Barlow)~\cite{fini2022self} && Dis & $62.7$ \\
Ours (CNN)           && $\times$   & \textbf{65.9 ± 0.7} \\
Ours (ViT)        && $\times$   & \textbf{64.5 ± 0.4} \\ 
\midrule
\multicolumn{4}{@{}l}{\textbf{CIFAR-100}} \\
\midrule
CLA (SimSiam)~\cite{pmlr-v330-cignoni26a}      & & M+Dis & $24.4 \pm 1.3$\\
Ours (CNN)       && $\times$   & 22.3 ± 4.1\\
Ours (ViT)            && $\times$   & \textbf{32.0 ± 1.1} \\
\bottomrule
\end{tabular}
\label{tab:unsup}
\end{table}

\begin{table*}[ht]
\centering
\caption{Ablation study on CIFAR-10 across 5 runs to evaluate the effect of each component on the average accuracy.
         \textbf{Frozen} = (\checkmark: frozen patches, $\times$: learnable patches), \textbf{Adapter} = (\checkmark with or $\times$ without adapter layers), \textbf{GFC} = Global Feature Centroid (\checkmark with or $\times$ without GFC).
         $\Delta$ vs full: change in accuracy relative to the full configuration (all-\checkmark).}
\setlength{\tabcolsep}{5pt}
\renewcommand{\arraystretch}{1.25}
\begin{tabular}{ccc cc cc}
\toprule
\multicolumn{3}{c}{\textbf{Configuration}} &
\multicolumn{2}{c}{\textbf{CNN}} &
\multicolumn{2}{c}{\textbf{ViT}} \\
\cmidrule(lr){1-3}\cmidrule(lr){4-5}\cmidrule(lr){6-7}
\textbf{Frozen} & \textbf{Adapter} & \textbf{GFC} &
\textbf{Acc.} & \textbf{$\Delta$ vs full} &
\textbf{Acc.} & \textbf{$\Delta$ vs full} \\
\midrule
\checkmark & \checkmark & \checkmark & \textbf{65.9 ± 0.7} & --- & \textbf{64.5 ± 0.4} & --- \\
$\times$   & \checkmark & \checkmark & 52.9 ± 2.4 & $-$13.0 & 61.1 ± 1.1 & $-$3.4 \\
\checkmark & $\times$   & $\times$   & 49.8 ± 3.2 & $-$16.1 & 61.7 ± 0.9 & $-$2.8 \\
\checkmark & \checkmark & $\times$   & 51.8 ± 1.6 & $-$14.1 & 61.0 ± 1.4 & $-$3.5 \\
$\times$   & $\times$   & $\times$   & 50.1 ± 2.8 & $-$15.8 & 61.6 ± 0.7 & $-$2.9 \\
\bottomrule
\end{tabular}
\label{tab:abl}
\end{table*}

\subsection{Ablation study}
\label{sec:ablation}

\subsubsection{Effect of the proposed components}
Table~\ref{tab:abl} reports an ablation study on CIFAR-10 across 5 runs by isolating the contribution of each architectural component including, frozen patch features, adapter layers, and the Global Feature Centroid (GFC) for both the CNN and ViT variants, relative to the full configuration with all three components enabled. In the \textbf{CNN setting}, the full configuration (frozen features + adapter + GFC) achieves the best result at $65.9$. Using trainable patches instead of frozen patch features while keeping the adapter and GFC active drops the accuracy to $52.9$ ($-13.0$). Disabling both the adapter and the GFC while keeping the frozen features yields $49.8$ ($-16.1$), which is the largest drop and even slightly below the all-disabled settings ($50.1$). The configuration with frozen 
features and adapter enabled but GFC disabled reaches $51.8$ ($-14.1$). 
Overall, every CNN ablation loses between $13$ and $16$ percentage relative to the full model. This indicates that for the CNN backbone the three components are strongly complementary. For the \textbf{ViT setting}, the pattern differs markedly. Removing any single component causes a much smaller drop than in the CNN setting: $61.1$ ($-3.4$) with learnable patches instead of frozen ones, $61.7$ ($-2.8$) without the adapter and the GFC, and $61.0$ ($-3.5$) without the GFC alone. The ViT ablations show that the frozen patch-level token representations already provide a strong foundation and that the ViT backbone is far more robust to component removal than the CNN backbone.

\subsubsection{Sensitivity to random seed}
\begin{figure*}
\centering
\includegraphics[width=1\linewidth]{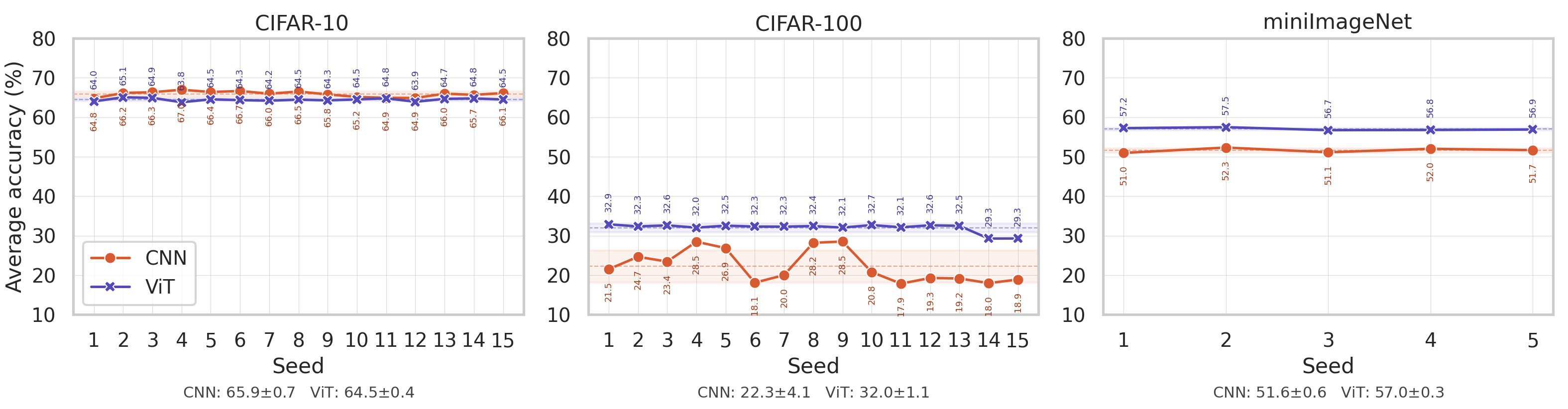}
\caption{Class incremental learning sensitivity to random seed on CIFAR-10 (5 splits), CIFAR-100 (20 splits), and  miniImageNet (20 splits).}
\label{fig:seed} 
\end{figure*}

Figure~\ref{fig:seed} shows the average accuracy over many random seeds (15 seeds for CIFAR-10 and CIFAR-100, 5 for miniImageNet) with the class order fixed. Table~\ref{tab:stats} lists the matching statistics. On CIFAR-10 the seed curves are almost flat. All 15 CNN seeds stay between $64.8$ and $67.0$, and all 15 ViT seeds between $63.8$ and $65.1$. The same is true on miniImageNet, where the five CNN seeds fall in $51.0$–$52.3$ and the five ViT seeds in $56.7$–$57.5$. In these cases the coefficient of variation is at most $1.2\%$ and the $95\%$ confidence interval is never wider than one point (for example $[65.5, 66.3]$ and $[64.3, 64.7]$ on CIFAR-10). This shows that the results are stable across random initializations. 

The one setting that behaves differently is the CNN variant on CIFAR-100. Its seed curve in Figure~\ref{fig:seed} is clearly unstable.  Some seeds reach $28$–$29$  while others drop to $18$–$21$, which produces the wide shaded band. This results lineup with the larger spread in Table~\ref{tab:stats} with  $18.4\%$ CV, $[20.0, 24.5]$ CI, and $[17.9, 28.5]$ range. Instead, the ViT variant on the same dataset is steadier: 13 of its 15 seeds are close to $32$–$33$, and only two drop to $29.3$, giving a small range of $[29.3, 32.9]$ and a CV of $3.5\%$. Comparing the two panels on the same dataset with the same number of tasks shows that the  ViT is the more stable backbone on every dataset, however, the CNN backbone is more seed-sensitive in case of larger number of splits.

\begin{table*}[ht]
\centering
\caption{Statistical robustness of average class incremental learning accuracy under two sources of randomness:
         random seed (repeated runs at fixed class order) and class order (random class-to-task
         permutations at fixed seed). \textbf{Runs} is the number of seeds or class orders.
         \textbf{CI}: 95\% confidence interval for the mean. \textbf{CV}: coefficient
         of variation.}
\setlength{\tabcolsep}{2.5pt}
\small
\begin{tabular}{@{}ll l r c c c c@{}}
\toprule
\textbf{} & \textbf{Dataset} & \textbf{Model} & \textbf{Runs} &
\textbf{Mean $\pm\,\sigma$ (\%)} & \textbf{95\% CI} & \textbf{CV} & \textbf{[min, max] (\%)} \\
\midrule
\multirow{6}{*}{\textbf{Seed}}
 & \multirow{2}{*}{CIFAR-10}     & CNN & 15 & $65.9 \pm 0.7$ & [65.5, 66.3] & 1.0  & [64.8, 67.0] \\
 &                               & ViT & 15 & $64.5 \pm 0.4$ & [64.3, 64.7] & 0.6  & [63.8, 65.1] \\
\cmidrule(l){2-8}
 & \multirow{2}{*}{CIFAR-100}    & CNN & 15 & $22.3 \pm 4.1$ & [20.0, 24.5] & 18.4 & [17.9, 28.5] \\
 &                               & ViT & 15 & $32.0 \pm 1.1$ & [31.4, 32.6] & 3.5  & [29.3, 32.9] \\
\cmidrule(l){2-8}
 & \multirow{2}{*}{miniImageNet} & CNN & 5  & $51.6 \pm 0.6$ & [50.9, 52.3] & 1.1  & [51.0, 52.3] \\
 &                               & ViT & 5  & $57.0 \pm 0.3$ & [56.6, 57.4] & 0.6  & [56.7, 57.5] \\
\midrule
\multirow{6}{*}{\shortstack[l]{\textbf{Class}\\\textbf{order}}}
 & \multirow{2}{*}{CIFAR-10}     & CNN & 5 & $67.1 \pm 2.6$ & [63.9, 70.3] & 3.8 & [62.9, 69.7] \\
 &                               & ViT & 5 & $65.7 \pm 2.4$ & [62.8, 68.6] & 3.6 & [61.8, 67.7] \\
\cmidrule(l){2-8}
 & \multirow{2}{*}{CIFAR-100}    & CNN & 5 & $30.5 \pm 1.5$ & [28.7, 32.3] & 4.8 & [28.8, 31.9] \\
 &                               & ViT & 5 & $35.4 \pm 1.0$ & [34.2, 36.7] & 2.8 & [33.9, 36.6] \\
\cmidrule(l){2-8}
 & \multirow{2}{*}{miniImageNet} & CNN & 5 & $20.5 \pm 3.8$ & [15.8, 25.2] & 18.4 & [14.2, 24.4] \\
 &                               & ViT & 5 & $25.0 \pm 0.9$ & [23.9, 26.1] & 3.5  & [24.0, 26.2] \\
\bottomrule
\end{tabular}
\label{tab:stats}
\end{table*}

\begin{figure*}
\centering
\includegraphics[width=1\linewidth]{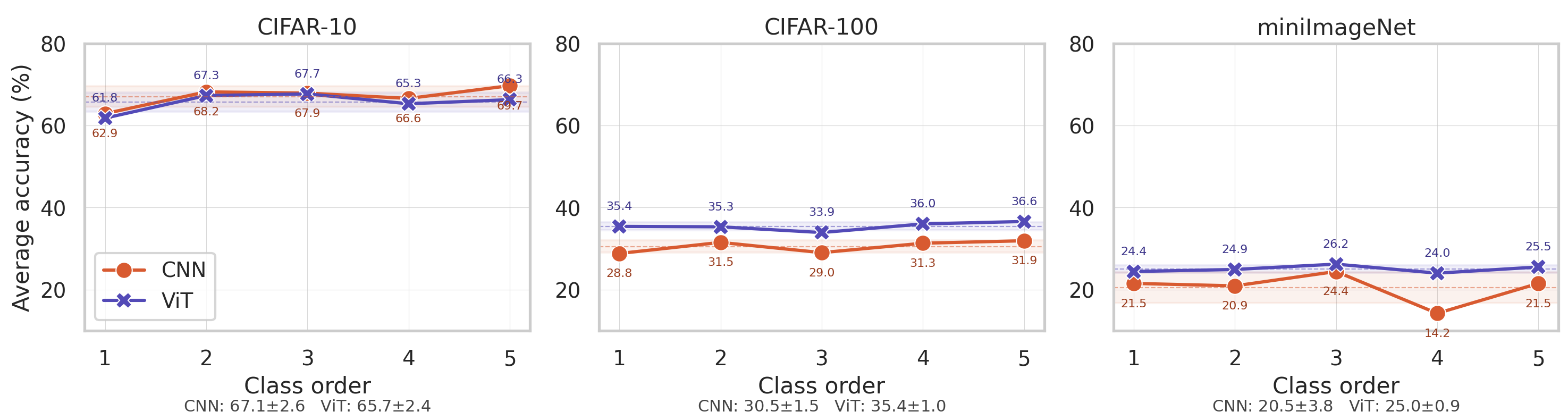}
\caption{Class incremental learning sensitivity to random class order permutation on CIFAR-10 (5 splits), CIFAR-100 (20 splits), and  miniImageNet (20 splits).}
\label{fig:order} 
\end{figure*}

\subsubsection{Sensitivity to class-order permutation}
Figure~\ref{fig:order} tests how much the order of classes affects accuracy, using five random class orders per setting with the seed fixed, and Table~\ref{tab:stats} provides the corresponding statistics. On CIFAR-10 the results of the five orders stay close together for both backbones, where CNN ranges from $62.9$ to $69.7$ and ViT from $61.8$ to $67.7$. On CIFAR-100 the per order results are even closer, with CNN in $28.8$–$31.9$ and ViT in $33.9$–$36.6$. In all four CIFAR cases the coefficient of variation is below $5\%$, so the accuracy does not depend on the specific class order. The two backbones differ most on miniImageNet. The five ViT orders are almost flat, covering only $24.0$–$26.2$ (a $2.2$-point range, CV $3.5\%$). The CNN curve is more unstable. Four of its five orders lie between $20.9$ and $24.4$, but one order drops to $14.2$. This single order widens the CNN statistics to $20.5 \pm 3.8$, CV $18.4\%$, CI $[15.8, 25.2]$. This is the same kind of instability the CNN variant showed for seeds on CIFAR-100, and again it appears only in the longer, harder setting. The ViT variant has no such outlier on any dataset. Together with the seed results, this shows that the method is robust to both random seeds and class order on the standard benchmarks, and that ViT is clearly more robust than CNN in the hardest settings, where CNN is still good on average but sometimes produces a low run.

\subsubsection{Cross-task effect of GFC.}
\begin{figure}
\centering
\includegraphics[width=1\linewidth]{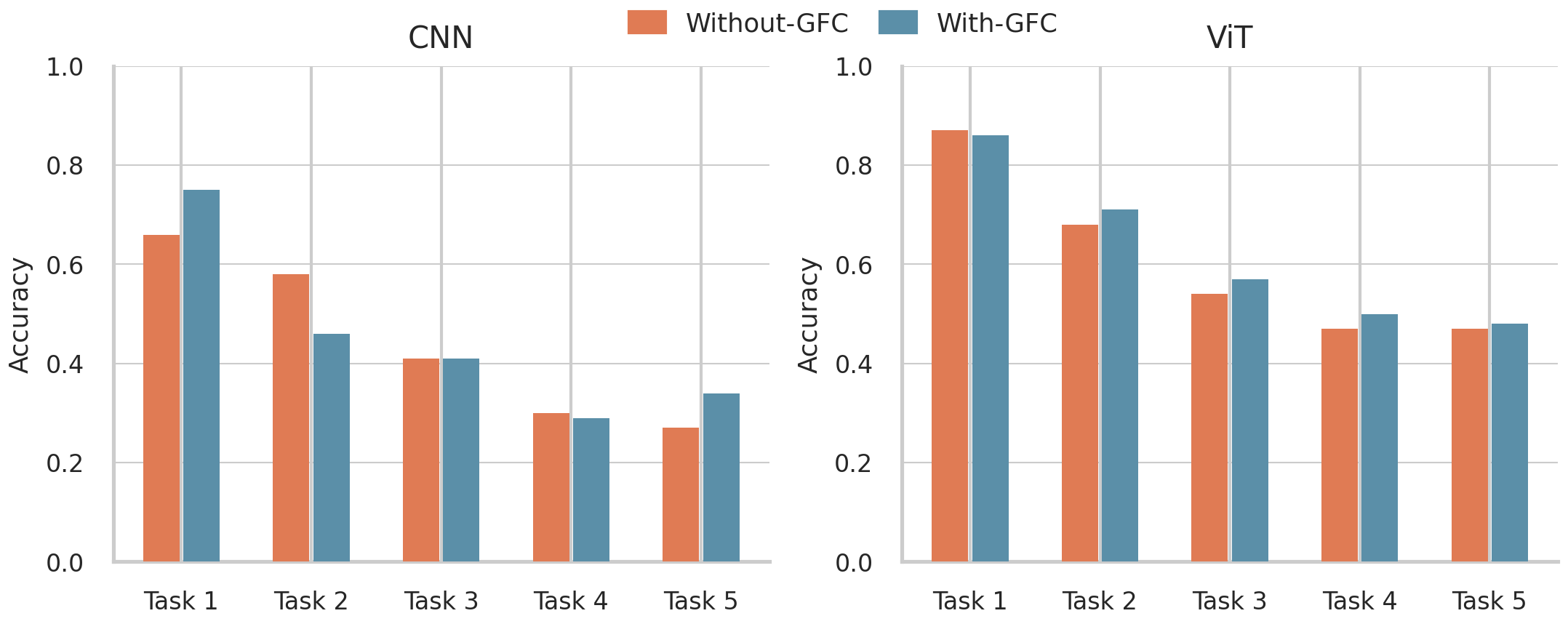}
\caption{Effect of the absence of GFC on the CIL accuracy, evaluated on CIFAR-10 with 5 tasks.}
\label{fig:gfm} 
\end{figure}

Figure~\ref{fig:gfm} plots per-task accuracy on CIFAR-10 with and without the GFC enabled, for both CNN and ViT. For \textbf{CNN}, the GFC helps most on the first and last tasks. The With-GFC bars are clearly higher on Task~1 and Task~5, roughly equal on Task~3, and slightly lower on Task~2 and Task~4. On average, enabling the GFC improves accuracy, which is consistent with the average value in ablation table. This can be attributed to that each new task's adapter update tends to overwrite prior representations, which the GFC mitigates. For  \textbf{ViT}, the two configurations remain close at every task, with the With-GFC bars holding only a small, consistent advantage over the Without-GFC 
bars, and no sharp decline occur. This is consistent with the ViT backbone's stability deriving primarily from its frozen patch-level tokens rather than the GFC. Overall, the GFC prevents an early, severe forgetting for CNN, while for ViT it provides only a smaller, steady accuracy improvement rather than acting as the primary component against forgetting.

\subsubsection{Effect of regularization weights $\lambda$ and $\alpha$.}
\begin{figure}
\centering
\includegraphics[width=1\linewidth]{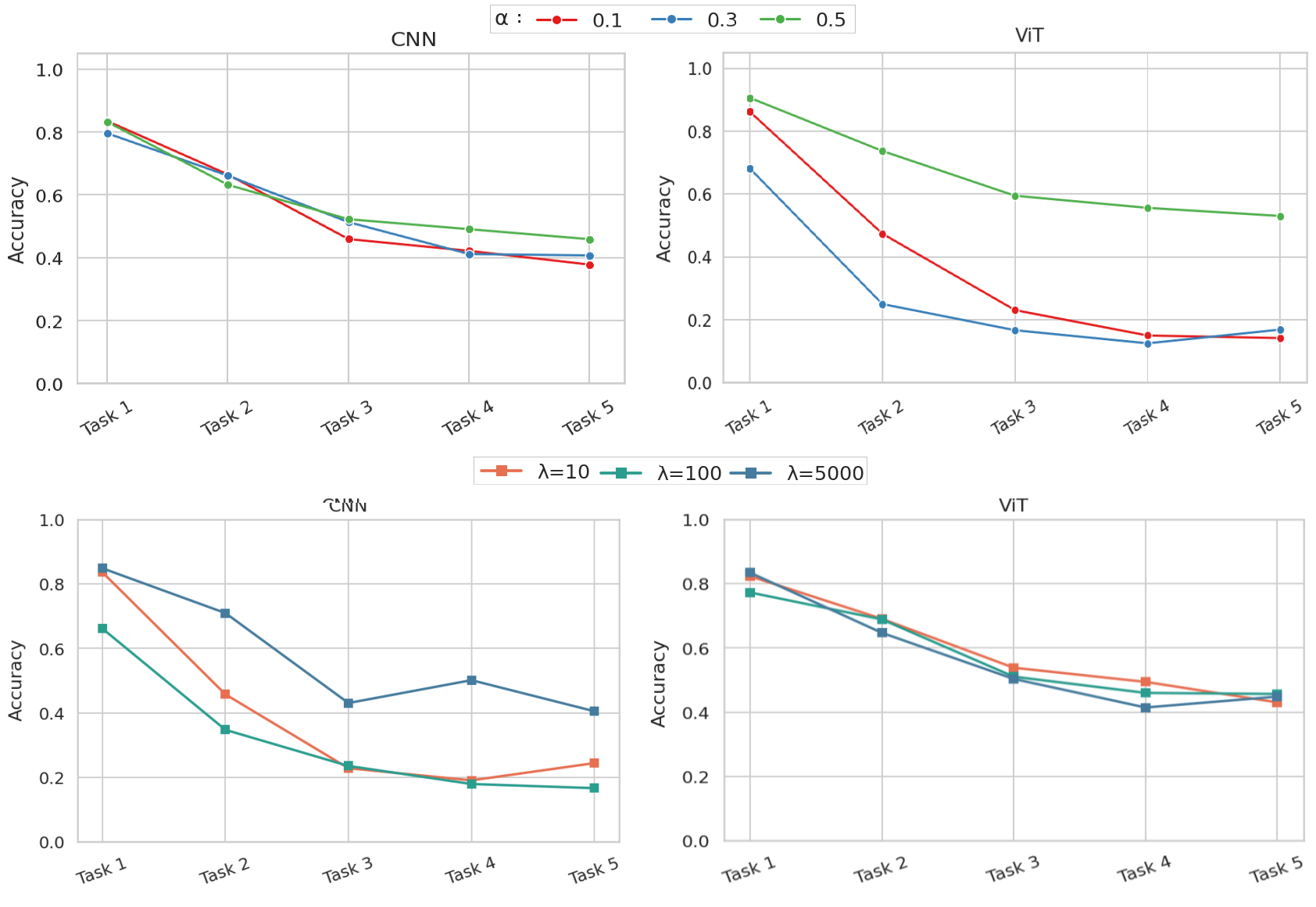}
\caption{Effect of the GFC weight $\alpha$ and the EWC weight $\lambda$ on CIL accuracy, evaluated on CIFAR-10 with 5 tasks.}
\label{fig:alpha_lambda} 
\end{figure}

Figure~\ref{fig:alpha_lambda} examines the sensitivity of PatchAdaNet to its two regularization weights, $\alpha$ (GFC) and $\lambda$ (EWC). In the top row, for \textbf{CNN}, all three values $\alpha \in \{0.1, 0.3, 0.5\}$ produce nearly overlapping accuracy trajectories across tasks, with only a slight edge for $\alpha=0.5$ at Task~5, which indicates that the CNN backbone is largely  insensitive to the exact value of $\alpha$. For \textbf{ViT}, the pattern is  different. With $\alpha=0.5$ maintains higher accuracy at  every task and declines only gradually, while $\alpha=0.1$ and $\alpha=0.3$ both decline after Task~1 and converge to much lower accuracy by Task~5. This indicates that for ViT, a stronger GFC weight is necessary to prevent early-task forgetting, in contrast to CNN where the GFC weight has little effect.

\begin{figure*}
\centering
\includegraphics[width=1\linewidth]{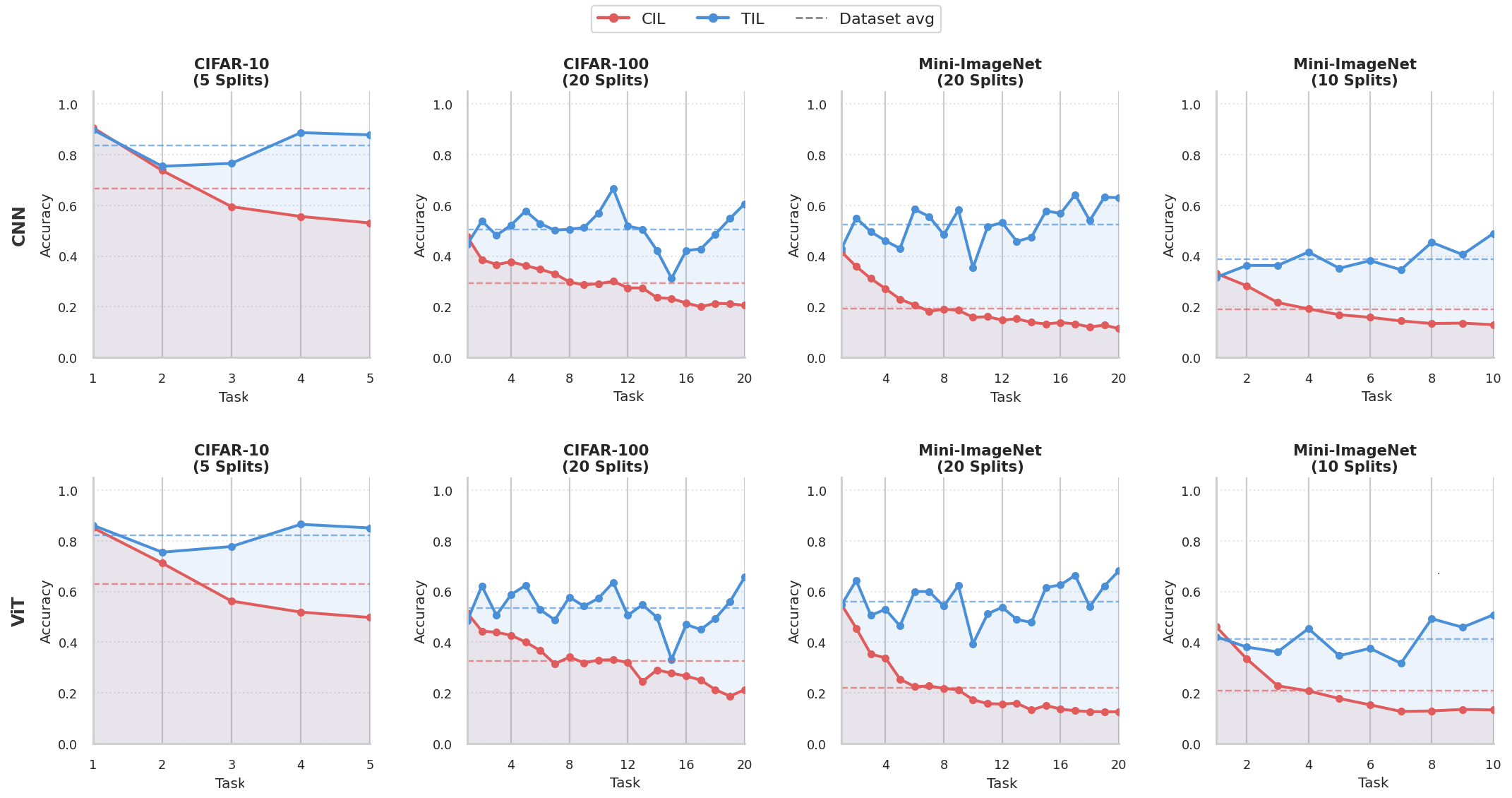}
\caption{Per task CIL and TIL accuracy on CIFAR-10, CIFAR-100, and miniImagenet.}
\label{fig:cil_til} 
\end{figure*}
In the bottom  row,  For \textbf{CNN}, the three settings $\lambda \in \{10, 100, 5000\}$ diverge considerably. $\lambda=5000$ achieves the most stable and highest accuracy across tasks, while $\lambda=10$ and $\lambda=100$ both decline more steeply. This  behavior suggests that too little protection against forgetting affect the accuracy in CNN settings. For \textbf{ViT}, all three $\lambda$ values track closely together throughout the task sequence, converging to similar accuracy by Task~5 regardless of the EWC weight used. This suggests that ViT's stability is 
comparatively insensitive to the EWC weighting, whereas the CNN backbone's adapter-driven representation is more sensitive to how strongly its weight-space updates.

\subsubsection{CIL and TIL across tasks}
Figure~\ref{fig:cil_til} compares Class-Incremental Learning (CIL) and Task-Incremental Learning (TIL) accuracy. Across every dataset and setting, TIL accuracy (blue) remains consistently and substantially higher than CIL accuracy (red) once more than two or three tasks have been seen, and the gap widens as the task sequence lengthens. On CIFAR-10, the two curves start close together at Task~1 (CNN: $\approx 93$ vs.\ $90$) but diverge from Task~3 onward, settling around $85$--$90$ for TIL versus $55$--$60$ for CIL by Task~5. The effect is most observed on the longer 20-task datasets, CIFAR-100 and the 20-split miniImageNet, where CIL accuracy declines steadily from roughly $50$ at Task~1 to $15$--$20$ by Task~20 for both settings, while TIL accuracy fluctuates around a comparatively stable band of $45$--$65$ throughout the full task sequence. On the 10-split miniImageNet, the same pattern noticed at a shorter, where CIL accuracy starts close to TIL at Task~1 ($\approx 33$--$42$) but falls steadily to around $13$--$14$ by Task~10, while TIL accuracy remains comparatively stable.

This gap is expected since TIL classifies only within each task's own classes, while CIL must discriminate the full set of classes seen so far without task identity, making the problem harder at every step. Still, CIL accuracy does not collapse outright across 20 tasks.  It declines only slightly and gradually, which suggests that feature representation remains comparatively stable even as the label space grows.

\begin{figure*}
\centering
\includegraphics[width=0.95\linewidth]{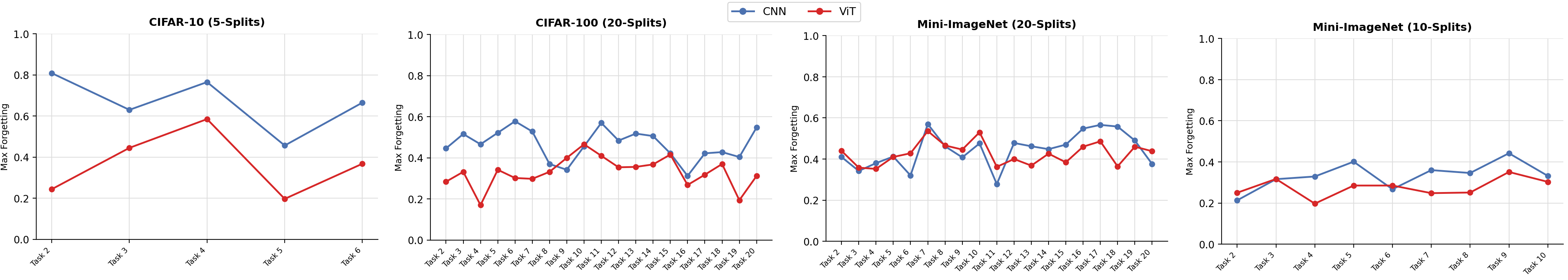}
\caption{Max forgetting after each task in CNN vs VIT settings.}
\label{fig:max_forg} 
\end{figure*}

\subsubsection{Forgetting Analysis}
\label{sec:forgetting}
Figure~\ref{fig:max_forg} reports the maximum forgetting metric after each task across the four benchmarks, comparing the CNN and ViT variants. Maximum forgetting at task $T_k$ is defined as the largest accuracy drop on any previously seen task's test set, relative to its accuracy when first evaluated, after training through $T_k$.

On \textbf{CIFAR-10}, Both CNN and ViT diverge sharply: CNN peaks at $\approx 81$ after Task~2 and stays above $60$ through Task~4 before dropping to $\approx 46$, while ViT starts lower ($\approx 24$) but rises to nearly converge with CNN ($58$) by Task~4, then falls to $19$. This is consistent with the earlier ablation finding that CNN is more forgetting-prone on this short 5-task sequence. On \textbf{CIFAR-100}, the gap is more persistent: CNN fluctuates between $30$--$60$ across all 20 tasks while ViT consistently tracks lower, mostly $20$--$45$, suggesting forgetting depends on which classes appear at each task rather than accumulating with task count.

On \textbf{miniImageNet}, the per task forgetting curves of CNN and ViT remain closely aligned throughout the task sequence, both fluctuating within the same $20$--$55$ range, and neither backbone exhibits a sustained advantage over the other. This indicates that, on this benchmark, the architectural differences between the frozen-feature reliance of CNN and ViT  do not translate into a measurable difference in forgetting.

\subsubsection{Task-order sensitivity across data domains}
\begin{figure*}
\centering
\includegraphics[width=1\linewidth]{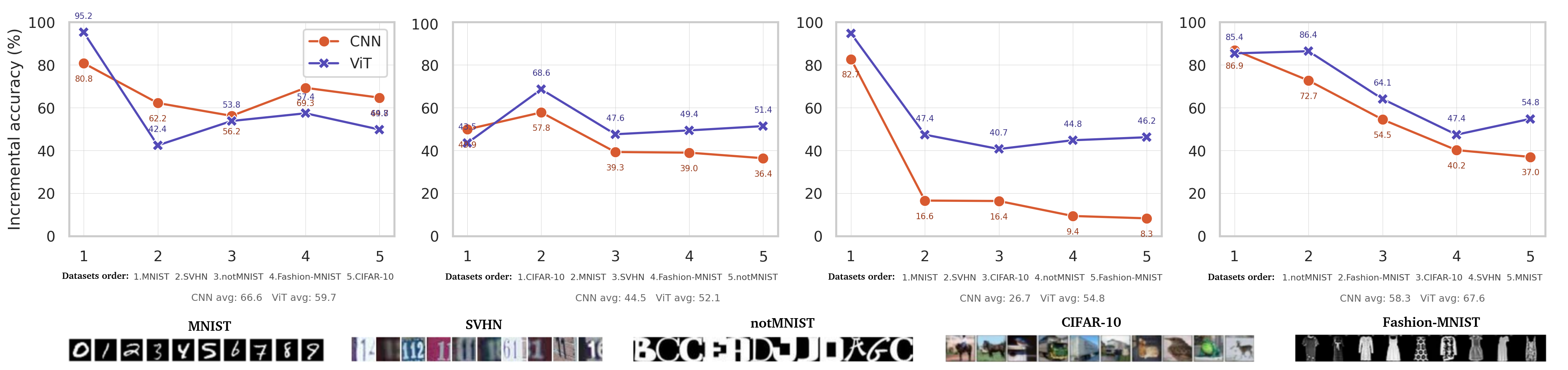}
\caption{Sensitivity of class-incremental accuracy to dataset order across heterogeneous dataset domains on the 5-Dataset sequence.}
\label{fig:5data} 
\end{figure*}

Figure~\ref{fig:5data} evaluates how sensitive CIL accuracy is to the order in which the five heterogeneous domains (MNIST, SVHN, notMNIST, Fashion-MNIST, CIFAR-10) are presented. Each plot corresponds to a different presentation order of the same five datasets, and reports the incremental accuracy after each task for both the CNN and ViT variants. Across the four orders, the average CIL accuracy varies widely for the CNN variant (from $26.7$ to $66.6$) but remains comparatively stable for the ViT variant (from $52.1$ to $67.6$). This indicates that the ViT backbone is substantially more robust to domain ordering than the CNN backbone.

The per-order curves reveal that the ordering effect is driven mainly by which domain is learned first and by specific hard transitions rather than by the task count. When a low-complexity grayscale dataset such as MNIST is placed first, both backbones start near their ceiling ($80.8$–$86.9$ for CNN, $85.4$–$95.2$ for ViT) but the subsequent shift to a very different domain (e.g., SVHN or CIFAR-10) causes a sharp drop, where the CNN variant is far more affected by these transitions, collapsing to $8.3$ by the final task in the MNIST$\rightarrow$SVHN$\rightarrow$CIFAR-10 order (Fig.~\ref{fig:5data}, third plot). The ViT variant absorbs the same transitions with a much shallower decline, retaining $46.2$ on the same order. Conversely, when the sequence begins with a harder domain such as notMNIST and interleaves domains more gradually (fourth plot), both variants retain higher accuracy throughout (CNN $58.3$, ViT $67.6$ on average). Overall, these results show that CIL accuracy under domain shift is governed  primarily by the difficulty of consecutive-task transitions and by the starting  domain, rather than by the number of tasks. The CNN variant, while competitive  under favorable orders, degrades severely when an easy domain is immediately  followed by dissimilar ones, whereas the ViT variant maintains a more stable  accuracy across all evaluated orders.

\subsection{Feature Visualization}
\label{sec:visualization}

\begin{figure*}
\centering
\includegraphics[width=0.9\linewidth]{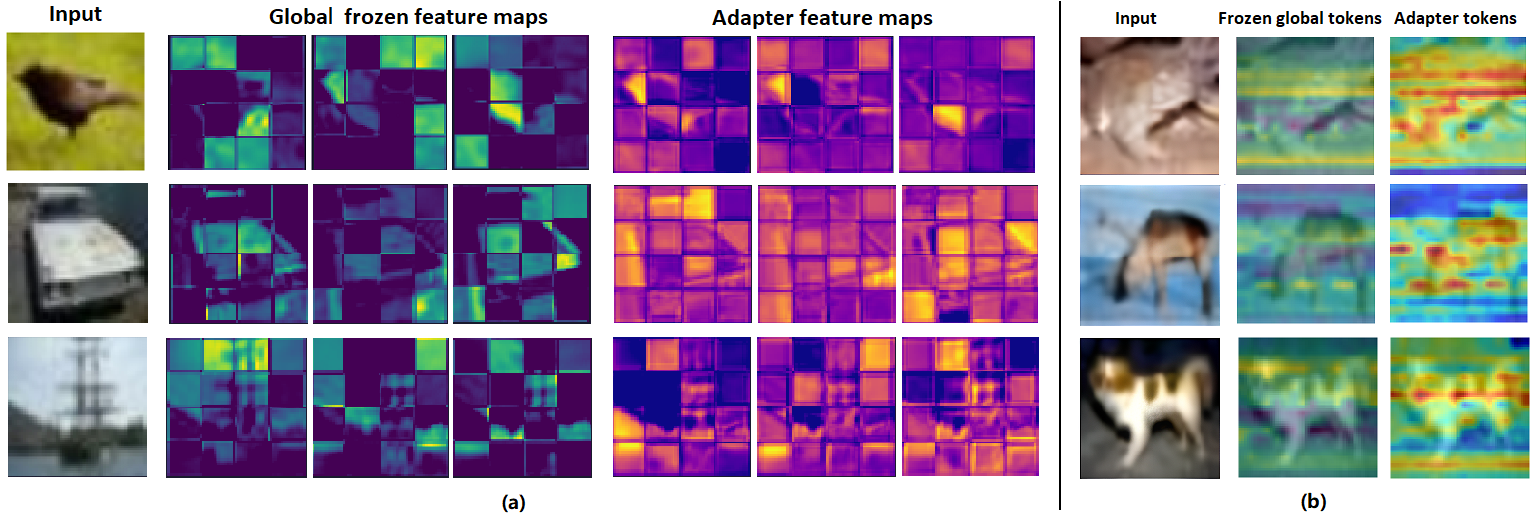}
\caption{Effect of local shared frozen features versus task-specific adapter representations. (a) CNN: tiled frozen CNN-PatchAE feature maps capture local, separated structure, while adapter layers in CNN-ImageAE learn global shapes. (b) ViT: frozen token activation maps focus on general-purpose local regions, while adapter token activation maps focus on task-specific object regions.}
\label{fig:maps} 
\end{figure*}

Figure~\ref{fig:maps} visualizes intermediate feature maps and activation maps reconstructed from token maps for sample inputs, separately for the CNN and ViT, respectively. The figure shows that frozen patch-level representations capture generic local structure shared across any task, while adapter layers refine this structure into task-specific representations.

\textbf{CNN feature maps:} The \emph{global frozen feature maps} 
$\mathbf{G}_{\mathrm{frozen}}^d$ show visible patch-grid boundaries with uniform, self-contained activation per tile, reflecting each PatchNet branch being pretrained in isolation with no access to neighboring patches or any downstream task; since the frozen weights are reused unchanged across all tasks, these patterns are generic rather than input-specific. The \emph{adapter feature maps} $\mathbf{G}_{\mathrm{adapt}}^d$ instead concentrate activation over object-relevant regions with smoother transitions across former patch boundaries, as the adapter learns which combinations of the fixed frozen patterns are diagnostic for the current task.

\textbf{ViT token maps:} Since ViT represents each patch as a token sequence, frozen and adapter tokens are mapped back to spatial coordinates for 
visualization. Because each frozen token is computed independently per patch with no cross-patch context, the reconstructed frozen token map is a 
combination of all patch-local activations and therefore spreads attention broadly and diffusely across the entire image, rather than concentrating on the salient object. Adapter token maps, by contrast, show attention reorganized into concentrated, object-specific regions that align closely with the input object's shape and boundaries, reflecting the cross-patch integration performed by the trainable fusion and adapter blocks.

\textbf{Interpretation:} Across both backbones, the frozen component encodes local structure that could in principle be found in any image from any task, while the adapter component reorganizes this shared structure into a task-specific representation. This division of labor allows the frozen parameters to remain immune to forgetting by construction, task-specific adaptation risk of catastrophic forgetting is alleviated using feature regularization (GFC) and weight regularization (EWC).

\section{Conclusion}
\label{sec:conclusion}
This paper presented PatchAdaNet, a replay-free, unsupervised class-incremental learning framework that separates reusable feature extraction from task-specific adaptation through a two-stage architecture. By pretraining independent patch-level autoencoders once on external data and freezing their encoder weights, PatchAdaNet establishes a feature foundation in which the frozen parameters do not receive gradient updates from any task-specific loss, so the shared representation is not subject to catastrophic forgetting. All adaptation across the task sequence is confined to a comparatively small set of learnable adapter layers, regularized jointly by the Global Feature Centroid (GFC) in feature space and Elastic Weight Consolidation (EWC) in weight space. Across the supervised and unsupervised evaluations, the method performed competitively with replay-based methods while storing no exemplars or past features.

\textbf{Limitations:} Several considerations remain open for the present work. The patch autoencoders and GFC were evaluated with relatively compact models with three-layer convolutional encoders for CNN and shallow Transformer blocks for ViT, pretrained on a comparatively small external dataset (TinyImageNet). It therefore remains unclear whether these low-level features generalize as well under greater visual complexity or domain diversity, particularly given the domain-shift sensitivity observed in the 5-Datasets evaluation. The GFC is also constructed from only the final mini-batch of each task rather than a full-dataset average, which may introduce sample-dependent noise. Finally, robustness varied across settings, where the CNN variant showed higher seed and class-order variance than the ViT variant on the longer CIFAR-100 and miniImageNet sequences, which indicates that stability is not uniform across backbones and benchmarks.

\textbf{Future work:} We aim to evaluate PatchAdaNet with larger backbone architectures and on larger-scale datasets to assess whether the benefits of the frozen patch-level feature foundation persist as representational capacity increases. In particular, the current lightweight patch autoencoders could be replaced with deeper or pretrained large-scale backbones for the patch-learning stage, leveraging large external pretraining datasets to learn richer and potentially more generalizable local structural representations before the continual adaptation phase begins. Whether scaling both the patch autoencoder and the continual-evaluation datasets improves accuracy and reduces forgetting remains an open question for future study.\\

\noindent



\bibliography{biblio}


\begin{thebibliography}{33}
\ifx \bisbn   \undefined \def \bisbn  #1{ISBN #1}\fi
\ifx \binits  \undefined \def \binits#1{#1}\fi
\ifx \bauthor  \undefined \def \bauthor#1{#1}\fi
\ifx \batitle  \undefined \def \batitle#1{#1}\fi
\ifx \bjtitle  \undefined \def \bjtitle#1{#1}\fi
\ifx \bvolume  \undefined \def \bvolume#1{\textbf{#1}}\fi
\ifx \byear  \undefined \def \byear#1{#1}\fi
\ifx \bissue  \undefined \def \bissue#1{#1}\fi
\ifx \bfpage  \undefined \def \bfpage#1{#1}\fi
\ifx \blpage  \undefined \def \blpage #1{#1}\fi
\ifx \burl  \undefined \def \burl#1{\textsf{#1}}\fi
\ifx \doiurl  \undefined \def \doiurl#1{\url{https://doi.org/#1}}\fi
\ifx \betal  \undefined \def \betal{\textit{et al.}}\fi
\ifx \binstitute  \undefined \def \binstitute#1{#1}\fi
\ifx \binstitutionaled  \undefined \def \binstitutionaled#1{#1}\fi
\ifx \bctitle  \undefined \def \bctitle#1{#1}\fi
\ifx \beditor  \undefined \def \beditor#1{#1}\fi
\ifx \bpublisher  \undefined \def \bpublisher#1{#1}\fi
\ifx \bbtitle  \undefined \def \bbtitle#1{#1}\fi
\ifx \bedition  \undefined \def \bedition#1{#1}\fi
\ifx \bseriesno  \undefined \def \bseriesno#1{#1}\fi
\ifx \blocation  \undefined \def \blocation#1{#1}\fi
\ifx \bsertitle  \undefined \def \bsertitle#1{#1}\fi
\ifx \bsnm \undefined \def \bsnm#1{#1}\fi
\ifx \bsuffix \undefined \def \bsuffix#1{#1}\fi
\ifx \bparticle \undefined \def \bparticle#1{#1}\fi
\ifx \barticle \undefined \def \barticle#1{#1}\fi
\bibcommenthead
\ifx \bconfdate \undefined \def \bconfdate #1{#1}\fi
\ifx \botherref \undefined \def \botherref #1{#1}\fi
\ifx \url \undefined \def \url#1{\textsf{#1}}\fi
\ifx \bchapter \undefined \def \bchapter#1{#1}\fi
\ifx \bbook \undefined \def \bbook#1{#1}\fi
\ifx \bcomment \undefined \def \bcomment#1{#1}\fi
\ifx \oauthor \undefined \def \oauthor#1{#1}\fi
\ifx \citeauthoryear \undefined \def \citeauthoryear#1{#1}\fi
\ifx \endbibitem  \undefined \def \endbibitem {}\fi
\ifx \bconflocation  \undefined \def \bconflocation#1{#1}\fi
\ifx \arxivurl  \undefined \def \arxivurl#1{\textsf{#1}}\fi
\csname PreBibitemsHook\endcsname

\bibitem[\protect\citeauthoryear{Kirkpatrick et~al.}{2017}]{kirkpatrick2017overcoming}
\begin{barticle}
\bauthor{\bsnm{Kirkpatrick}, \binits{J.}},
\bauthor{\bsnm{Pascanu}, \binits{R.}},
\bauthor{\bsnm{Rabinowitz}, \binits{N.}},
\bauthor{\bsnm{Veness}, \binits{J.}},
\bauthor{\bsnm{Desjardins}, \binits{G.}},
\bauthor{\bsnm{Rusu}, \binits{A.A.}},
\bauthor{\bsnm{Milan}, \binits{K.}},
\bauthor{\bsnm{Quan}, \binits{J.}},
\bauthor{\bsnm{Ramalho}, \binits{T.}},
\bauthor{\bsnm{Grabska-Barwinska}, \binits{A.}}, \betal:
\batitle{Overcoming catastrophic forgetting in neural networks}.
\bjtitle{Proceedings of the national academy of sciences}
\bvolume{114}(\bissue{13}),
\bfpage{3521}--\blpage{3526}
(\byear{2017})
\end{barticle}
\endbibitem

\bibitem[\protect\citeauthoryear{Zenke et~al.}{2017}]{zenke2017continual}
\begin{bchapter}
\bauthor{\bsnm{Zenke}, \binits{F.}},
\bauthor{\bsnm{Poole}, \binits{B.}},
\bauthor{\bsnm{Ganguli}, \binits{S.}}:
\bctitle{Continual learning through synaptic intelligence}.
In: \bbtitle{International Conference on Machine Learning},
pp. \bfpage{3987}--\blpage{3995}
(\byear{2017}).
\bcomment{PMLR}
\end{bchapter}
\endbibitem

\bibitem[\protect\citeauthoryear{Aljundi et~al.}{2018}]{aljundi2018memory}
\begin{bchapter}
\bauthor{\bsnm{Aljundi}, \binits{R.}},
\bauthor{\bsnm{Babiloni}, \binits{F.}},
\bauthor{\bsnm{Elhoseiny}, \binits{M.}},
\bauthor{\bsnm{Rohrbach}, \binits{M.}},
\bauthor{\bsnm{Tuytelaars}, \binits{T.}}:
\bctitle{Memory aware synapses: Learning what (not) to forget}.
In: \bbtitle{Proceedings of the European Conference on Computer Vision (ECCV)},
pp. \bfpage{139}--\blpage{154}
(\byear{2018})
\end{bchapter}
\endbibitem

\bibitem[\protect\citeauthoryear{Rusu et~al.}{2016}]{rusu2016progressive}
\begin{botherref}
\oauthor{\bsnm{Rusu}, \binits{A.A.}},
\oauthor{\bsnm{Rabinowitz}, \binits{N.C.}},
\oauthor{\bsnm{Desjardins}, \binits{G.}},
\oauthor{\bsnm{Soyer}, \binits{H.}},
\oauthor{\bsnm{Kirkpatrick}, \binits{J.}},
\oauthor{\bsnm{Kavukcuoglu}, \binits{K.}},
\oauthor{\bsnm{Pascanu}, \binits{R.}},
\oauthor{\bsnm{Hadsell}, \binits{R.}}:
Progressive neural networks.
arXiv preprint arXiv:1606.04671
(2016)
\end{botherref}
\endbibitem

\bibitem[\protect\citeauthoryear{Yoon et~al.}{2018}]{yoon2018lifelong}
\begin{bchapter}
\bauthor{\bsnm{Yoon}, \binits{J.}},
\bauthor{\bsnm{Yang}, \binits{E.}},
\bauthor{\bsnm{Lee}, \binits{J.}},
\bauthor{\bsnm{Hwang}, \binits{S.J.}}:
\bctitle{Lifelong learning with dynamically expandable networks}.
In: \bbtitle{International Conference on Learning Representations}
(\byear{2018})
\end{bchapter}
\endbibitem

\bibitem[\protect\citeauthoryear{Lopez-Paz and Ranzato}{2017}]{lopez2017gradient}
\begin{botherref}
\oauthor{\bsnm{Lopez-Paz}, \binits{D.}},
\oauthor{\bsnm{Ranzato}, \binits{M.}}:
Gradient episodic memory for continual learning.
Advances in neural information processing systems
\textbf{30}
(2017)
\end{botherref}
\endbibitem

\bibitem[\protect\citeauthoryear{Shin et~al.}{2017}]{shin2017continual}
\begin{botherref}
\oauthor{\bsnm{Shin}, \binits{H.}},
\oauthor{\bsnm{Lee}, \binits{J.K.}},
\oauthor{\bsnm{Kim}, \binits{J.}},
\oauthor{\bsnm{Kim}, \binits{J.}}:
Continual learning with deep generative replay.
Advances in neural information processing systems
\textbf{30}
(2017)
\end{botherref}
\endbibitem

\bibitem[\protect\citeauthoryear{Li and Hoiem}{2017}]{li2017learning}
\begin{barticle}
\bauthor{\bsnm{Li}, \binits{Z.}},
\bauthor{\bsnm{Hoiem}, \binits{D.}}:
\batitle{Learning without forgetting}.
\bjtitle{IEEE transactions on pattern analysis and machine intelligence}
\bvolume{40}(\bissue{12}),
\bfpage{2935}--\blpage{2947}
(\byear{2017})
\end{barticle}
\endbibitem

\bibitem[\protect\citeauthoryear{Wen et~al.}{2024}]{wen2024class}
\begin{bchapter}
\bauthor{\bsnm{Wen}, \binits{H.}},
\bauthor{\bsnm{Pan}, \binits{L.}},
\bauthor{\bsnm{Dai}, \binits{Y.}},
\bauthor{\bsnm{Qiu}, \binits{H.}},
\bauthor{\bsnm{Wang}, \binits{L.}},
\bauthor{\bsnm{Wu}, \binits{Q.}},
\bauthor{\bsnm{Li}, \binits{H.}}:
\bctitle{Class incremental learning with multi-teacher distillation}.
In: \bbtitle{Proceedings of the IEEE/CVF Conference on Computer Vision and Pattern Recognition},
pp. \bfpage{28443}--\blpage{28452}
(\byear{2024})
\end{bchapter}
\endbibitem

\bibitem[\protect\citeauthoryear{Ahn et~al.}{2019}]{ahn2019uncertainty}
\begin{botherref}
\oauthor{\bsnm{Ahn}, \binits{H.}},
\oauthor{\bsnm{Cha}, \binits{S.}},
\oauthor{\bsnm{Lee}, \binits{D.}},
\oauthor{\bsnm{Moon}, \binits{T.}}:
Uncertainty-based continual learning with adaptive regularization.
Advances in neural information processing systems
\textbf{32}
(2019)
\end{botherref}
\endbibitem

\bibitem[\protect\citeauthoryear{Chen et~al.}{2021}]{chen2021overcoming}
\begin{bchapter}
\bauthor{\bsnm{Chen}, \binits{P.-H.}},
\bauthor{\bsnm{Wei}, \binits{W.}},
\bauthor{\bsnm{Hsieh}, \binits{C.-J.}},
\bauthor{\bsnm{Dai}, \binits{B.}}:
\bctitle{Overcoming catastrophic forgetting by bayesian generative regularization}.
In: \bbtitle{International Conference on Machine Learning},
pp. \bfpage{1760}--\blpage{1770}
(\byear{2021}).
\bcomment{PMLR}
\end{bchapter}
\endbibitem

\bibitem[\protect\citeauthoryear{Ebrahimi et~al.}{2020}]{ebrahimi2020adversarial}
\begin{bchapter}
\bauthor{\bsnm{Ebrahimi}, \binits{S.}},
\bauthor{\bsnm{Meier}, \binits{F.}},
\bauthor{\bsnm{Calandra}, \binits{R.}},
\bauthor{\bsnm{Darrell}, \binits{T.}},
\bauthor{\bsnm{Rohrbach}, \binits{M.}}:
\bctitle{Adversarial continual learning}.
In: \bbtitle{European Conference on Computer Vision},
pp. \bfpage{386}--\blpage{402}
(\byear{2020}).
\bcomment{Springer}
\end{bchapter}
\endbibitem

\bibitem[\protect\citeauthoryear{Kozal and Wozniak}{2023}]{kozal2023increasing}
\begin{barticle}
\bauthor{\bsnm{Kozal}, \binits{J.}},
\bauthor{\bsnm{Wozniak}, \binits{M.}}:
\batitle{Increasing depth of neural networks for life-long learning}.
\bjtitle{Information Fusion}
\bvolume{98},
\bfpage{101829}
(\byear{2023})
\end{barticle}
\endbibitem

\bibitem[\protect\citeauthoryear{Zbontar et~al.}{2021}]{zbontar2021barlow}
\begin{bchapter}
\bauthor{\bsnm{Zbontar}, \binits{J.}},
\bauthor{\bsnm{Jing}, \binits{L.}},
\bauthor{\bsnm{Misra}, \binits{I.}},
\bauthor{\bsnm{LeCun}, \binits{Y.}},
\bauthor{\bsnm{Deny}, \binits{S.}}:
\bctitle{Barlow twins: Self-supervised learning via redundancy reduction}.
In: \bbtitle{International Conference on Machine Learning},
pp. \bfpage{12310}--\blpage{12320}
(\byear{2021}).
\bcomment{PMLR}
\end{bchapter}
\endbibitem

\bibitem[\protect\citeauthoryear{Chen et~al.}{2020}]{chen2020simple}
\begin{bchapter}
\bauthor{\bsnm{Chen}, \binits{T.}},
\bauthor{\bsnm{Kornblith}, \binits{S.}},
\bauthor{\bsnm{Norouzi}, \binits{M.}},
\bauthor{\bsnm{Hinton}, \binits{G.}}:
\bctitle{A simple framework for contrastive learning of visual representations}.
In: \bbtitle{International Conference on Machine Learning},
pp. \bfpage{1597}--\blpage{1607}
(\byear{2020}).
\bcomment{PmLR}
\end{bchapter}
\endbibitem

\bibitem[\protect\citeauthoryear{Cha et~al.}{2021}]{cha2021co2l}
\begin{bchapter}
\bauthor{\bsnm{Cha}, \binits{H.}},
\bauthor{\bsnm{Lee}, \binits{J.}},
\bauthor{\bsnm{Shin}, \binits{J.}}:
\bctitle{Co2l: Contrastive continual learning}.
In: \bbtitle{Proceedings of the IEEE/CVF International Conference on Computer Vision},
pp. \bfpage{9516}--\blpage{9525}
(\byear{2021})
\end{bchapter}
\endbibitem

\bibitem[\protect\citeauthoryear{Fini et~al.}{2022}]{fini2022self}
\begin{bchapter}
\bauthor{\bsnm{Fini}, \binits{E.}},
\bauthor{\bsnm{Da~Costa}, \binits{V.G.T.}},
\bauthor{\bsnm{Alameda-Pineda}, \binits{X.}},
\bauthor{\bsnm{Ricci}, \binits{E.}},
\bauthor{\bsnm{Alahari}, \binits{K.}},
\bauthor{\bsnm{Mairal}, \binits{J.}}:
\bctitle{Self-supervised models are continual learners}.
In: \bbtitle{Proceedings of the IEEE/CVF Conference on Computer Vision and Pattern Recognition},
pp. \bfpage{9621}--\blpage{9630}
(\byear{2022})
\end{bchapter}
\endbibitem

\bibitem[\protect\citeauthoryear{Wen et~al.}{2024}]{wen2024provable}
\begin{bchapter}
\bauthor{\bsnm{Wen}, \binits{Y.}},
\bauthor{\bsnm{Tan}, \binits{Z.}},
\bauthor{\bsnm{Zheng}, \binits{K.}},
\bauthor{\bsnm{Xie}, \binits{C.}},
\bauthor{\bsnm{Huang}, \binits{W.}}:
\bctitle{Provable contrastive continual learning}.
In: \bbtitle{International Conference on Machine Learning},
pp. \bfpage{52819}--\blpage{52838}
(\byear{2024}).
\bcomment{PMLR}
\end{bchapter}
\endbibitem

\bibitem[\protect\citeauthoryear{Cheng et~al.}{2023}]{cheng2023cucl}
\begin{bchapter}
\bauthor{\bsnm{Cheng}, \binits{C.}},
\bauthor{\bsnm{Song}, \binits{J.}},
\bauthor{\bsnm{Zhu}, \binits{X.}},
\bauthor{\bsnm{Zhu}, \binits{J.}},
\bauthor{\bsnm{Gao}, \binits{L.}},
\bauthor{\bsnm{Shen}, \binits{H.}}:
\bctitle{Cucl: Codebook for unsupervised continual learning}.
In: \bbtitle{Proceedings of the 31st ACM International Conference on Multimedia},
pp. \bfpage{1729}--\blpage{1737}
(\byear{2023})
\end{bchapter}
\endbibitem

\bibitem[\protect\citeauthoryear{Cheng et~al.}{2024}]{cheng2024must}
\begin{bchapter}
\bauthor{\bsnm{Cheng}, \binits{H.}},
\bauthor{\bsnm{Wen}, \binits{H.}},
\bauthor{\bsnm{Qiu}, \binits{H.}},
\bauthor{\bsnm{Wang}, \binits{L.}},
\bauthor{\bsnm{Zhang}, \binits{M.}},
\bauthor{\bsnm{Li}, \binits{H.}}:
\bctitle{Must unsupervised continual learning relies on previous information?}
In: \bbtitle{Proceedings of the IEEE/CVF Conference on Computer Vision and Pattern Recognition},
pp. \bfpage{5519}--\blpage{5529}
(\byear{2024})
\end{bchapter}
\endbibitem

\bibitem[\protect\citeauthoryear{Chaudhry et~al.}{}]{chaudhryefficient}
\begin{botherref}
\oauthor{\bsnm{Chaudhry}, \binits{A.}},
\oauthor{\bsnm{Ranzato}, \binits{M.}},
\oauthor{\bsnm{Rohrbach}, \binits{M.}},
\oauthor{\bsnm{Elhoseiny}, \binits{M.}}:
Efficient lifelong learning with a-gem.
In: International Conference on Learning Representations
\end{botherref}
\endbibitem

\bibitem[\protect\citeauthoryear{Aljundi et~al.}{2019a}]{aljundi2019gradient}
\begin{botherref}
\oauthor{\bsnm{Aljundi}, \binits{R.}},
\oauthor{\bsnm{Lin}, \binits{M.}},
\oauthor{\bsnm{Goujaud}, \binits{B.}},
\oauthor{\bsnm{Bengio}, \binits{Y.}}:
Gradient based sample selection for online continual learning.
Advances in neural information processing systems
\textbf{32}
(2019)
\end{botherref}
\endbibitem

\bibitem[\protect\citeauthoryear{Aljundi et~al.}{2019b}]{aljundi2019online}
\begin{botherref}
\oauthor{\bsnm{Aljundi}, \binits{R.}},
\oauthor{\bsnm{Belilovsky}, \binits{E.}},
\oauthor{\bsnm{Tuytelaars}, \binits{T.}},
\oauthor{\bsnm{Charlin}, \binits{L.}},
\oauthor{\bsnm{Caccia}, \binits{M.}},
\oauthor{\bsnm{Lin}, \binits{M.}},
\oauthor{\bsnm{Page-Caccia}, \binits{L.}}:
Online continual learning with maximal interfered retrieval.
Advances in neural information processing systems
\textbf{32}
(2019)
\end{botherref}
\endbibitem

\bibitem[\protect\citeauthoryear{Prabhu et~al.}{2020}]{prabhu2020gdumb}
\begin{bchapter}
\bauthor{\bsnm{Prabhu}, \binits{A.}},
\bauthor{\bsnm{Torr}, \binits{P.H.}},
\bauthor{\bsnm{Dokania}, \binits{P.K.}}:
\bctitle{Gdumb: A simple approach that questions our progress in continual learning}.
In: \bbtitle{European Conference on Computer Vision},
pp. \bfpage{524}--\blpage{540}
(\byear{2020}).
\bcomment{Springer}
\end{bchapter}
\endbibitem

\bibitem[\protect\citeauthoryear{Shim et~al.}{2021}]{shim2021online}
\begin{bchapter}
\bauthor{\bsnm{Shim}, \binits{D.}},
\bauthor{\bsnm{Mai}, \binits{Z.}},
\bauthor{\bsnm{Jeong}, \binits{J.}},
\bauthor{\bsnm{Sanner}, \binits{S.}},
\bauthor{\bsnm{Kim}, \binits{H.}},
\bauthor{\bsnm{Jang}, \binits{J.}}:
\bctitle{Online class-incremental continual learning with adversarial shapley value}.
In: \bbtitle{Proceedings of the AAAI Conference on Artificial Intelligence},
vol. \bseriesno{35},
pp. \bfpage{9630}--\blpage{9638}
(\byear{2021})
\end{bchapter}
\endbibitem

\bibitem[\protect\citeauthoryear{Mai et~al.}{2021}]{mai2021supervised}
\begin{bchapter}
\bauthor{\bsnm{Mai}, \binits{Z.}},
\bauthor{\bsnm{Li}, \binits{R.}},
\bauthor{\bsnm{Kim}, \binits{H.}},
\bauthor{\bsnm{Sanner}, \binits{S.}}:
\bctitle{Supervised contrastive replay: Revisiting the nearest class mean classifier in online class-incremental continual learning}.
In: \bbtitle{Proceedings of the IEEE/CVF Conference on Computer Vision and Pattern Recognition},
pp. \bfpage{3589}--\blpage{3599}
(\byear{2021})
\end{bchapter}
\endbibitem

\bibitem[\protect\citeauthoryear{De~Lange and Tuytelaars}{2021}]{de2021continual}
\begin{bchapter}
\bauthor{\bsnm{De~Lange}, \binits{M.}},
\bauthor{\bsnm{Tuytelaars}, \binits{T.}}:
\bctitle{Continual prototype evolution: Learning online from non-stationary data streams}.
In: \bbtitle{Proceedings of the IEEE/CVF International Conference on Computer Vision},
pp. \bfpage{8250}--\blpage{8259}
(\byear{2021})
\end{bchapter}
\endbibitem

\bibitem[\protect\citeauthoryear{Gu et~al.}{2022}]{gu2022not}
\begin{bchapter}
\bauthor{\bsnm{Gu}, \binits{Y.}},
\bauthor{\bsnm{Yang}, \binits{X.}},
\bauthor{\bsnm{Wei}, \binits{K.}},
\bauthor{\bsnm{Deng}, \binits{C.}}:
\bctitle{Not just selection, but exploration: Online class-incremental continual learning via dual view consistency}.
In: \bbtitle{Proceedings of the IEEE/CVF Conference on Computer Vision and Pattern Recognition},
pp. \bfpage{7442}--\blpage{7451}
(\byear{2022})
\end{bchapter}
\endbibitem

\bibitem[\protect\citeauthoryear{Guo et~al.}{2022}]{guo2022online}
\begin{bchapter}
\bauthor{\bsnm{Guo}, \binits{Y.}},
\bauthor{\bsnm{Liu}, \binits{B.}},
\bauthor{\bsnm{Zhao}, \binits{D.}}:
\bctitle{Online continual learning through mutual information maximization}.
In: \bbtitle{International Conference on Machine Learning},
pp. \bfpage{8109}--\blpage{8126}
(\byear{2022}).
\bcomment{PMLR}
\end{bchapter}
\endbibitem

\bibitem[\protect\citeauthoryear{Wei et~al.}{2023}]{wei2023online}
\begin{bchapter}
\bauthor{\bsnm{Wei}, \binits{Y.}},
\bauthor{\bsnm{Ye}, \binits{J.}},
\bauthor{\bsnm{Huang}, \binits{Z.}},
\bauthor{\bsnm{Zhang}, \binits{J.}},
\bauthor{\bsnm{Shan}, \binits{H.}}:
\bctitle{Online prototype learning for online continual learning}.
In: \bbtitle{Proceedings of the IEEE/CVF International Conference on Computer Vision},
pp. \bfpage{18764}--\blpage{18774}
(\byear{2023})
\end{bchapter}
\endbibitem

\bibitem[\protect\citeauthoryear{Derakhshani et~al.}{2021}]{derakhshani2021kernel}
\begin{bchapter}
\bauthor{\bsnm{Derakhshani}, \binits{M.M.}},
\bauthor{\bsnm{Zhen}, \binits{X.}},
\bauthor{\bsnm{Shao}, \binits{L.}},
\bauthor{\bsnm{Snoek}, \binits{C.}}:
\bctitle{Kernel continual learning}.
In: \bbtitle{International Conference on Machine Learning},
pp. \bfpage{2621}--\blpage{2631}
(\byear{2021}).
\bcomment{PMLR}
\end{bchapter}
\endbibitem

\bibitem[\protect\citeauthoryear{Cignoni et~al.}{2026}]{pmlr-v330-cignoni26a}
\begin{bchapter}
\bauthor{\bsnm{Cignoni}, \binits{G.}},
\bauthor{\bsnm{Cossu}, \binits{A.}},
\bauthor{\bsnm{Gomez-Villa}, \binits{A.}},
\bauthor{\bsnm{Weijer}, \binits{J.v.d.}},
\bauthor{\bsnm{Carta}, \binits{A.}}:
\bctitle{Cla: Latent alignment for online continual self-supervised learning}.
In: \bbtitle{Proceedings of The 4th Conference on Lifelong Learning Agents}.
\bsertitle{Proceedings of Machine Learning Research},
pp. \bfpage{756}--\blpage{775}
(\byear{2026})
\end{bchapter}
\endbibitem

\bibitem[\protect\citeauthoryear{Mirzadeh et~al.}{2020}]{mirzadeh2020understanding}
\begin{barticle}
\bauthor{\bsnm{Mirzadeh}, \binits{S.I.}},
\bauthor{\bsnm{Farajtabar}, \binits{M.}},
\bauthor{\bsnm{Pascanu}, \binits{R.}},
\bauthor{\bsnm{Ghasemzadeh}, \binits{H.}}:
\batitle{Understanding the role of training regimes in continual learning}.
\bjtitle{Advances in neural information processing systems}
\bvolume{33},
\bfpage{7308}--\blpage{7320}
(\byear{2020})
\end{barticle}
\endbibitem

\end{thebibliography}

\end{document}